\pdfoutput=1
\documentclass{article}

\PassOptionsToPackage{table}{xcolor}
\usepackage[table]{xcolor}
\usepackage{amsmath}
\usepackage{amssymb}
\usepackage{amsfonts}
\usepackage{graphicx}
\usepackage{multirow}
\usepackage{booktabs}
\usepackage{caption}
\usepackage{enumitem}
\usepackage{tabularx}
\usepackage{subcaption}
\usepackage{tcolorbox}
\tcbuselibrary{breakable}
\usepackage[accsupp]{axessibility}

\usepackage[preprint]{corl_2026}

\definecolor{cosmosblue}{RGB}{230,243,255}
\definecolor{grootgreen}{RGB}{233,248,240}

\title{\textsc{World2Act}: Latent Action Post-Training from World Model Dynamics}

\author{
  An Dinh Vuong\thanks{Equal contribution}\\
  MBZUAI\\
  \texttt{an.vuong@mbzuai.ac.ae}
  \And
  Tuan Van Vo\protect\footnotemark[1]\\
  MBZUAI
  \And
  Abdullah Sohail\\
  MBZUAI
  \And
  Haoran Ding\\
  MBZUAI
  \And
  Liang Ma\\
  MBZUAI
  \And
  Xiaodan Liang\\
  MBZUAI
  \And
  Anqing Duan\\
  MBZUAI
  \And
  Ivan Laptev\\
  MBZUAI
  \And
  Ian Reid\\
  MBZUAI
}

\begin{document}

\maketitle

\begin{abstract}
    World Models (WMs) offer a promising mechanism for post-training Vision-Language-Action (VLA) policies by providing dynamics priors that improve generalization under task and scene variation. However, most WM-based post-training methods rely on pixel-space supervision, making policies sensitive to visual artifacts introduced by imperfect WM rollouts. We present \textsc{World2Act}, a latent-space post-training framework that transfers WM dynamics to the VLA policy without pixel-space supervision. \textsc{World2Act} operates in two stages: \textit{(i)} it induces a shared video--action latent space by contrastively aligning WM-dynamics latents with action embeddings, and \textit{(ii)} it post-trains the VLA by guiding policy action representations toward WM-imagined dynamics rather than decoded pixels. Built on GR00T-N1.6, \textsc{World2Act} delivers absolute success-rate gains of up to +2.5\% on simulation benchmarks (RoboCasa, LIBERO, Bridge-SIMPLER) and +6.7\% on a real robot over finetuned VLA baselines. Notably, it outperforms pixel-space WM supervision by up to +6.0\%, including on LIBERO where pixel supervision \textit{degrades} the baseline, suggesting that latent WM dynamics offer a more stable WM-based post-training alternative to pixel-space transfer.
\end{abstract}


\section{Introduction}
\label{sec:intro}
\begin{figure}[!ht]
    \centering
    \vspace{-2ex}
    \begin{subfigure}[t]{0.365\linewidth}
        \centering
        \includegraphics[width=\linewidth]{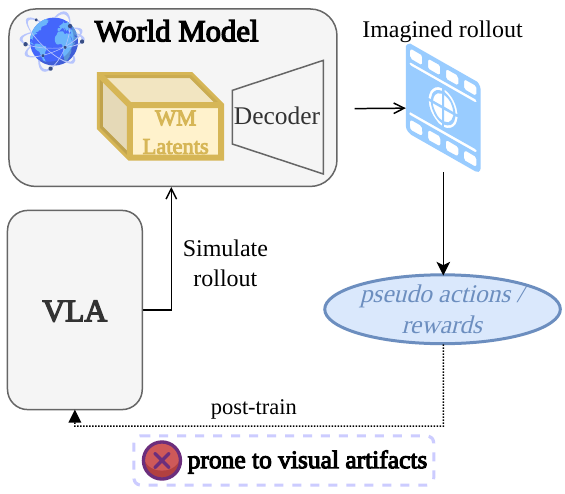}
        \vspace{-3ex}
        \caption{Pixel-space post-training.}
        \label{fig:teaser-pixel}
    \end{subfigure}\hfill
    \begin{subfigure}[t]{0.312\linewidth}
        \centering
        \includegraphics[width=\linewidth]{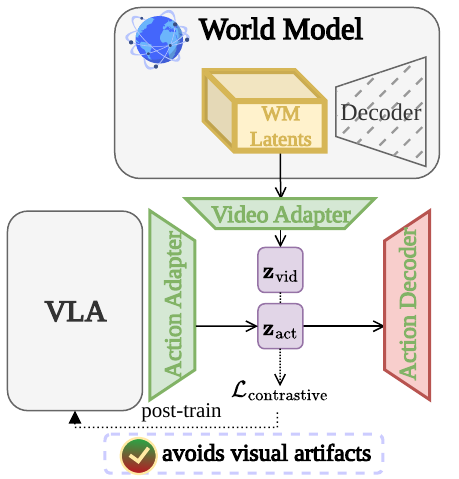}
         \vspace{-3ex}
        \caption{Latent post-training (Ours).}
        \label{fig:teaser-ours}
    \end{subfigure}\hfill
    \begin{subfigure}[t]{0.291\linewidth}
        \centering
        \includegraphics[width=\linewidth]{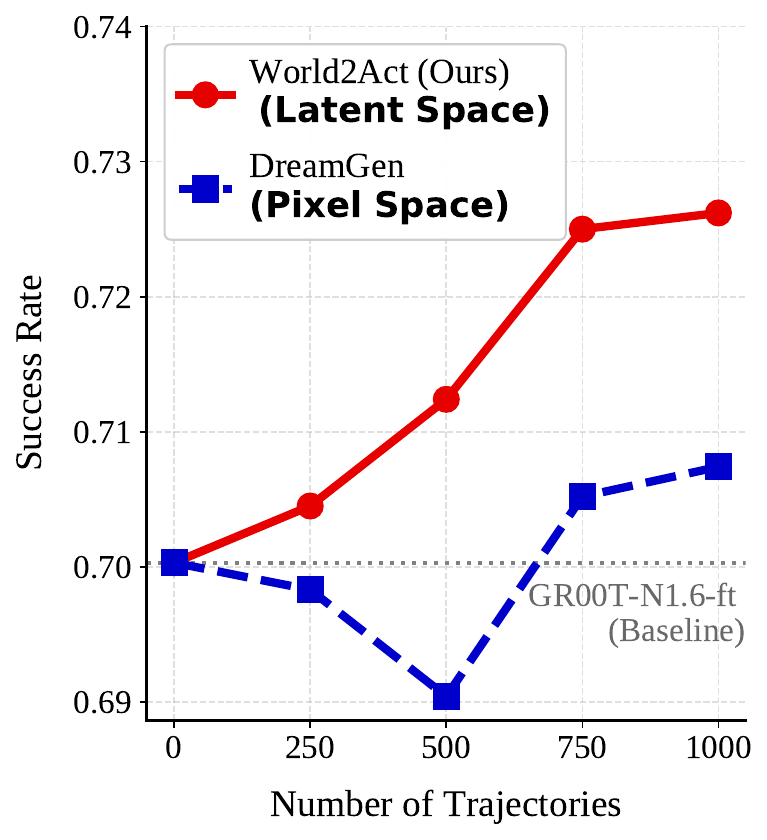}
         \vspace{-3ex}
        \caption{Post-training performance.}
        \label{fig:scaling-trajectories}
    \end{subfigure}
    \vspace{-1ex}
    \caption{\textit{(a) Problem:} Prior WM$\rightarrow$VLA post-training relies on imagined pixel rollouts, making policies vulnerable to visual artifacts. \textit{(b) Solution:} We bypass this by directly aligning VLA action representation with WM latent space. \textit{(c)} Post-training on GR00T-N1.6~\citep{nvidia_gear_gr00t_n1_6_2025} indicates that our latent post-training yields consistent improvements, overcoming the instability of pixel-space methods.}
    \label{fig:teaser}
    \vspace{-2ex}
\end{figure}

World models (WMs) offer a promising source of dynamics priors for embodied agents by modeling how robot-object interactions evolve over time~\citep{wang2026contactgaussian, kim2026cosmos}. In contrast, most vision-language-action (VLA) policies are learned primarily via behavior cloning and can struggle with environmental changes or unseen contact conditions, even after finetuning on extensive datasets~\citep{zhang2025bridging, reuss2025state-vla-iclr26}. This generalization gap suggests that VLAs lack robust dynamics priors that finetuning alone cannot provide~\citep{tan2025interactive}, motivating post-training methods that inject these priors from WMs~\citep{li2025vla}.

Previous WM-based post-training work~\citep{li2025vla, xiao2025world} largely targets pixel-to-action supervision via an inverse dynamics model (IDM)~\citep{jang2025dreamgen}, or reward signals~\citep{guo2025ctrl}. However, because these signals are computed from WM rollouts, their quality is limited by the fidelity of the generated pixels. When rollouts contain visual artifacts~\citep{poppi2026countervid}, or inconsistent contacts~\citep{shang2026roboscape}, these errors can be directly converted into noisy action labels or rewards, and may compound over long horizons~\citep{liu2025worldweaver}. Such noisy training signals can harm VLA policy performance, necessitating more reliable post-training approaches.

Video latents from WMs capture transferable behavioral dynamics~\citep{gao2025adaworld}, offering a natural alternative to pixels for transferring knowledge from WMs to VLAs~\citep{assran2025v}. Prior latent-video-to-action methods typically learn unified video--action representations, such as UWM~\citep{zhu2025unified} and Cosmos Policy~\citep{kim2026cosmos}. However, these frameworks often rely on high-dimensional joint embedding spaces, which can destabilize training~\citep{guo2025improving}. We instead post-train the VLA by matching policy action latents to WM-predicted latent dynamics targets in a shared low-dimensional video--action latent space. This bypasses pixel-space action refinement and reduces sensitivity to visual artifacts in WM rollouts.

We therefore introduce \textsc{World2Act}, a latent post-training method that aligns VLA action representation with WM latent video dynamics. Rather than supervising the policy from imagined pixels, \textsc{World2Act} uses WM latent trajectories as targets for policy action latents. Our method proceeds in two stages. First, we train video and action adapters that project WM video latents and expert action sequences into a shared representation space using a contrastive matching objective. Second, we post-train GR00T-N1.6~\citep{nvidia_gear_gr00t_n1_6_2025} by regularizing its action features toward this shared latent space, leaving the VLA backbone unchanged. Across RoboCasa~\citep{nasiriany2024robocasa}, LIBERO~\citep{liu2023libero}, Bridge-SIMPLER~\citep{zhou2025autoeval}, and real-world robot evaluations, \textsc{World2Act} improves post-training performance over pixel-space supervision, as illustrated in Fig.~\ref{fig:teaser}. In summary, our contributions are threefold:

\begin{itemize}[leftmargin=*, labelsep=0.5em]
    \vspace{-1.5ex}
  \item We identify sensitivity to visual artifacts in imagined rollouts as a key limitation of pixel-space WM post-training, motivating latent-space dynamics alignment as a more reliable alternative.
  \item We introduce \textsc{World2Act}, a two-stage post-training method that aligns VLA action representations with WM video-dynamics latents through a shared video--action adapter space.
  \item With GR00T-N1.6 as the VLA backbone, \textsc{World2Act} consistently outperforms pixel-space post-training methods and other baselines on simulation~\citep{nasiriany2024robocasa, liu2023libero, zhou2025autoeval} and real-world benchmarks.
   \vspace{-1.5ex}
\end{itemize}


\section{Related Work}
\label{sec: related_work}
\vspace{-1ex}

\noindent\textbf{World Models for Policy Improvement.} WMs have emerged as a key component in developing generalist robots~\citep{gao2026dreamdojo}. Because real-world data collection is serial, slow, and labor-intensive, WMs provide a low-cost, resettable, and highly scalable alternative for simulating environments and finetuning policies~\citep{yang2026rise}. Prior work leveraging WMs to improve policies can be broadly grouped into two categories~\citep{guo2026flowdreamer}: \textit{i)} \textit{Model-based control}, where WMs act as virtual environments to support methods such as model-based planning~\citep{jain2025smooth, zheng2025flare, ye2025latent, li2025vla}. These approaches rely on interactive, action-conditioned WMs, such as V-JEPA 2~\citep{assran2025v}, to produce imagined rollouts that guide the policy toward preferred trajectories~\citep{guo2026vlaw}. However, training such interactive models and subsequently rolling out numerous imagined trajectories can be computationally expensive and difficult to parallelize with policy optimization~\citep{guo2025ctrl}. \textit{ii)} \textit{Instruction-based control}, which conditions generation on task specifications, such as natural-language descriptions, to produce complete sequences of future states or actions~\citep{huangenerverse, zhou2024robodreamer, jang2025dreamgen, du2023learning}. This work adopts an instruction-based paradigm that enables simultaneous video generation and policy optimization in an efficient, parallel post-training pipeline.

\noindent\textbf{Post-Training Vision-Language-Action Models.} VLA backbones are commonly pretrained on web-scale instruction-following data and then finetuned on task- or robot-specific datasets~\citep{guo2026vlaw}. However, internet data primarily captures static semantics and broad vision–language knowledge, whereas robotics data is narrow, temporally structured, and action-centric; this mismatch can degrade downstream control performance~\citep{reuss2025state-vla-iclr26}. Recent work, therefore, moves beyond web-data finetuning~\citep{ye2025latent, bu2025univla, chen2025moto, zheng2025universal, zhu2025unified} toward WM post-training, which can provide dynamics and action-relevant structure~\citep{guo2025ctrl}. Existing methods, however, mainly focus on pixel space: they infer pseudo-actions from generated videos using IDMs~\citep{jang2025dreamgen} or learn rewards from pixel rollouts~\citep{li2025vla, jang2025dreamgen, xiao2025world, jiang2025irl, sharma2026world, guo2025ctrl}. Because generated rollouts can exhibit visual artifacts, these pixel-derived signals may introduce noisy supervision~\citep{poppi2026countervid} and degrade post-training. We address this limitation by aligning video and action representations directly in a shared video--action latent space.

\noindent\textbf{Latent Actions.}
Latent action representations have been explored as effective interfaces for learning from offline robot data and unlabeled videos across embodiments~\citep{ye2025latent, gao2025adaworld, bu2025univla, he2026pre, zhang2026latent}. More recent methods further learn joint latent video--action spaces that couple world modeling with action prediction, shifting generic video modeling toward action-aware dynamics~\citep{zhu2025unified, kim2026cosmos, yang2026chain}. While these methods bridge visual dynamics and action prediction, they often rely on high-dimensional joint embeddings that introduce architectural overhead when adapting existing VLA backbones. We instead align WM dynamics latents with VLA action features through lightweight adapters for post-training.
\vspace{-0.5ex}
\section{Method}
\label{sec: method}
\vspace{-1.5ex}
We present \textsc{World2Act}, a post-training method that transfers visual-dynamic priors from a frozen instruction-conditioned video WM to a VLA policy, as shown in Fig.~\ref{fig:method_overview}. We first describe the WM latent representation, then present our two-stage pipeline: \textit{i) Stage 1}: learning a shared video--action latent space that aligns WM latent dynamics with robot actions via bridge adapters, and \textit{ii) Stage 2}: training a lightweight residual policy on top of a frozen VLA, guided by the aligned latent space.

\vspace{-1.7ex}
\subsection{Preliminaries: Instruction-Conditioned Video World Model}
\label{subsec:vwm}
\vspace{-1ex}

We instantiate the world model \(\mathcal{W}\) with Cosmos-Predict2~\citep{agarwal2025cosmos}, an instruction-conditioned video WM for future visual-state prediction that operates in a structured video-VAE latent space. During pretraining, ground-truth videos are encoded by the video VAE into latent trajectories \(\mathbf{V}^{\mathrm{gt}}=\{V^{\mathrm{gt}}_t\}_{t=1}^{T}\), and a denoising network \(\mathcal{W}_{\mathrm{denoise}}\) is trained via flow matching~\citep{blattmann2023stable} to predict the vector field that transports noisy latents toward \(\mathbf{V}^{\mathrm{gt}}\) in this structured latent space. Given the initial observation \(s_1\) and language instruction \(\ell\), \(\mathcal{W}_{\mathrm{denoise}}\) predicts future visual dynamics in the video-VAE latent space:
\begin{equation}
    \label{eq:wm-denoise}
    \mathbf{V} = \mathcal{W}_{\mathrm{denoise}}(\epsilon; s_1, \ell),
    \qquad
    \mathbf{V}=\{V_t\}_{t=1}^{T}, \qquad V_t\in\mathbb{R}^{C\times H\times W}
\end{equation}
where \(\epsilon\) is the initial noise and each \(V_t\) is a denoised spatial latent corresponding to a non-overlapping chunk of \(M\) frames. In \textsc{World2Act}, we freeze all components of \(\mathcal{W}\), preserving its structured video-VAE latent space, and use video latents \(\mathbf{V}\) before VAE decoding as compact visual-dynamics priors. Since these latents describe what should visually happen but do not specify robot controls, we train a video--action bridge to map them into an action-aware latent space for VLA post-training.

\vspace{-1.5ex}
\subsection{Stage 1: Aligning Video Dynamics and Robot Actions}\label{sub-sec:Stage1}
\vspace{-1ex}

Stage 1 learns a video--action latent bridge from synchronized state-action demonstrations. Given an expert trajectory video, the frozen world model \(\mathcal{W}\) obtains a video latent sequence \(\mathbf{V}=\{V_t\}_{t=1}^{T}\), with each \(V_t\) corresponding to a chunk of \(M\) low-level steps, as described in~\autoref{subsec:vwm}.

\begin{figure*}[!ht]
    \centering
    \vspace{-3ex}
    \includegraphics[width=\linewidth, trim={2.15cm 0 6.8cm 0}, clip]{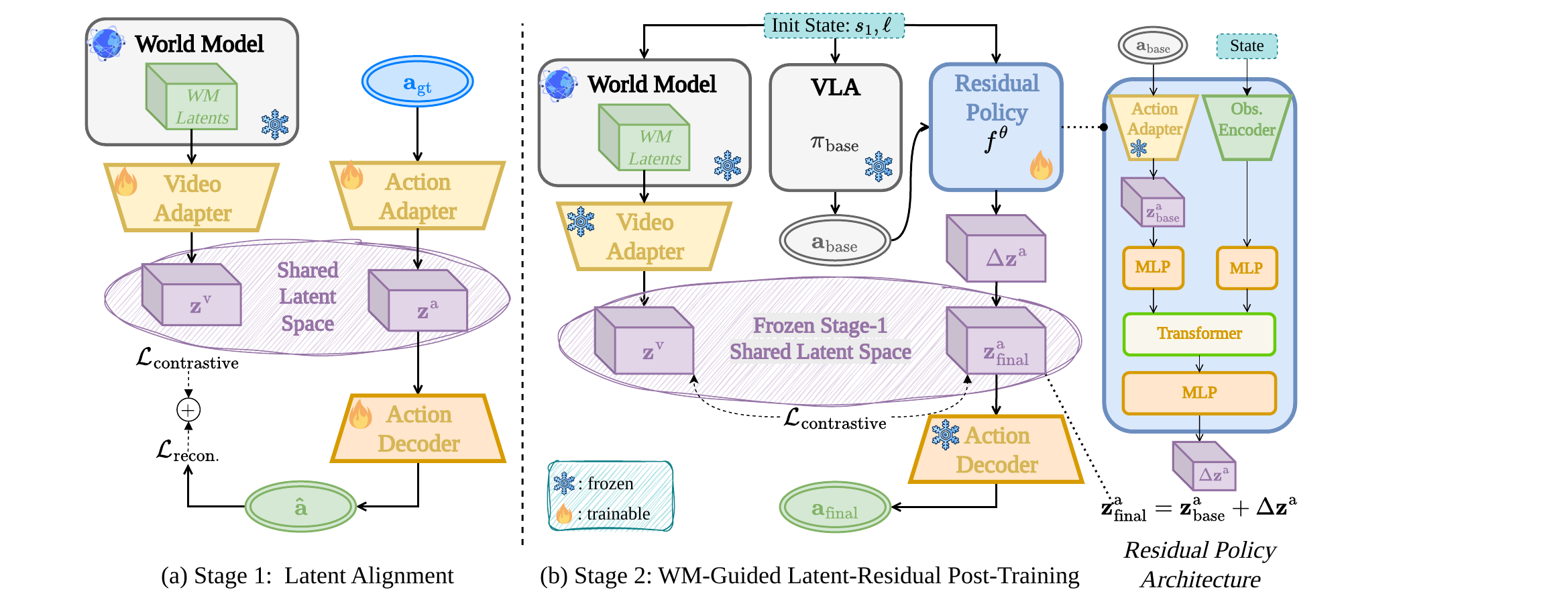}
    \vspace{-3ex}
    \caption{\textbf{\textsc{World2Act} overview.} \textbf{(a)} \textit{Stage 1: Latent alignment.} We train video and action adapters ($\mathcal{B}_\text{v},\mathcal{B}_\text{a}$) with reconstruction and contrastive objectives. \textbf{(b)} \textit{Stage 2: VLA post-training.} We freeze the VLA and learn a residual policy guided by WM-induced latent dynamics.}
    \label{fig:method_overview}
    \vspace{-3ex}
\end{figure*}

A CNN-based \textit{Video Adapter} \(\mathcal{B}_{\mathrm{v}}\) maps each WM latent to a \(D\)-dimensional embedding, producing \(\mathbf{z}^{\mathrm{v}}=\{z^{\mathrm{v}}_t\}_{t=1}^{T}\in\mathbb{R}^{T\times D}\), where \(z^{\mathrm{v}}_t=\mathcal{B}_{\mathrm{v}}(V_t)\). In parallel, an MLP-based \textit{Action Adapter} \(\mathcal{B}_{\mathrm{a}}\) partitions \(\mathbf{a}_{\mathrm{gt}}\) into temporally aligned action chunks \(\bar{a}^{\mathrm{gt}}_t=[a^{\mathrm{gt}}_{(t-1)M+1};\ldots;a^{\mathrm{gt}}_{tM}]\) and maps each chunk to \(z^{\mathrm{a}}_t=\mathcal{B}_{\mathrm{a}}(\bar{a}^{\mathrm{gt}}_t)\), yielding \(\mathbf{z}^{\mathrm{a}}=\{z^{\mathrm{a}}_t\}_{t=1}^{T}\in\mathbb{R}^{T\times D}\). This produces temporally aligned video and action embeddings while retaining within-chunk control variations, as depicted in Fig.~\ref{fig:method_overview}a.

To ensure the action latents preserve kinematics and temporal structure, we introduce an MLP-based \textit{Action Decoder} \(\mathcal{D}_{\mathrm{a}}\), which maps each action chunk latent \(z^{\mathrm{a}}_t\) back to \(M\) low-level action vectors. We train \(\mathcal{B}_{\mathrm{v}}\), \(\mathcal{B}_{\mathrm{a}}\), and \(\mathcal{D}_{\mathrm{a}}\) with two objectives. First, we use a reconstruction objective: unrolling the decoded chunks gives the reconstructed action sequence \(\hat{\mathbf{a}}\), supervised by
\(\mathcal{L}_{\mathrm{recon}}=\|\mathbf{a}_{\mathrm{gt}}-\hat{\mathbf{a}}\|^2\).

Second, we align video and action dynamics with a bidirectional InfoNCE loss~\citep{oord2018representation}, which encourages paired video--action trajectories to be close in the shared latent space while separating pairs from different demonstrations. A key design choice is whether to contrast a single global trajectory embedding or to preserve temporal correspondence across the trajectory. We adopt a chunk-aware trajectory alignment by computing similarity as the average cosine similarity over temporally aligned chunks. This reduces shortcut solutions where the model matches sequences using coarse task identity or background cues rather than fine-grained temporal dynamics. For each positive pair \((\mathbf{z}^{\mathrm{v}}_i,\mathbf{z}^{\mathrm{a}}_i)\) from the same demonstration, the remaining \(B{-}1\) samples in the batch serve as negatives, including \textit{i)} \emph{easy} negatives from different tasks and \textit{ii)} \emph{hard} negatives from other demonstrations of the same task. For a paired video--action sample \(i\) in a batch of size \(B\), the loss is defined as:
\begin{equation}
    \label{eq:alignment_loss}
    \mathcal{L}_{\mathrm{contrastive}}
    =
    - \log
    \frac{
    \exp(\mathrm{sim}(\mathbf{z}^{\mathrm{v}}_i,\mathbf{z}^{\mathrm{a}}_i)/\tau)
    }{
    \sum_{j=1}^{B}
    \exp(\mathrm{sim}(\mathbf{z}^{\mathrm{v}}_i,\mathbf{z}^{\mathrm{a}}_j)/\tau)
    }
    -
    \log
    \frac{
    \exp(\mathrm{sim}(\mathbf{z}^{\mathrm{a}}_i,\mathbf{z}^{\mathrm{v}}_i)/\tau)
    }{
    \sum_{j=1}^{B}
    \exp(\mathrm{sim}(\mathbf{z}^{\mathrm{a}}_i,\mathbf{z}^{\mathrm{v}}_j)/\tau)
    },
\end{equation}
where \(\tau\) is a temperature and
\(\mathrm{sim}(\mathbf{z}^{\mathrm{v}}_i,\mathbf{z}^{\mathrm{a}}_j)
=
\frac{1}{T}\sum_{t=1}^{T}
\cos(z^{\mathrm{v}}_{i,t},z^{\mathrm{a}}_{j,t})\)
denotes chunk-averaged cosine similarity. We set \(\tau{=}0.1\)~\citep{chen2020simple} and optimize
\(\mathcal{L}=\mathcal{L}_{\mathrm{recon}}+\mathcal{L}_{\mathrm{contrastive}}\).
After Stage 1, \(\mathcal{B}_{\mathrm{v}}\), \(\mathcal{B}_{\mathrm{a}}\), and \(\mathcal{D}_{\mathrm{a}}\) are frozen to preserve a shared latent space between video dynamics and decodable robot actions.

\vspace{-1.7ex}
\subsection{Stage 2: World-Model-Guided Latent-Residual Post-Training}\label{sub-sec:Stage-2}
\vspace{-1ex}

Inspired by Silver \textit{et al.}~\citep{silver2018residual}, we avoid post-training the heavily parameterized VLA policy, which can be sample-inefficient and prone to catastrophic forgetting~\citep{reuss2025state-vla-iclr26}. Instead, we freeze the VLA backbone \(\pi_{\mathrm{base}}\) and train a lightweight \textit{Residual Policy} \(f^\theta\) that predicts an additive correction in the learned action-latent space, as in Fig.~\ref{fig:method_overview}b. To train this residual policy, we use Stage-1-aligned WM latents as weak supervision for action-latent corrections, since the WM encodes imagined task dynamics rather than executable control labels. For a single rollout, given state \(s_t\) and instruction \(\ell\), the base VLA predicts an action chunk \(\bar{a}_{\mathrm{base},t}= \pi_{\mathrm{base}}(s_t, \ell)\), which is mapped to the action-latent space. \(f^\theta\) then refines this latent by combining the base action latent with a learned residual latent:
\begin{equation}\label{eq:latent_residual_policy}
z^{\mathrm a}_{\mathrm{base},t}
= \mathcal{B}_{\mathrm a}(\bar{a}_{\mathrm{base},t}), \qquad
\Delta z^{\mathrm a}_t
= f^\theta(s_t,z^{\mathrm a}_{\mathrm{base},t}), \qquad
z^{\mathrm a}_{\mathrm{final},t}
= z^{\mathrm a}_{\mathrm{base},t}+\Delta z^{\mathrm a}_t .
\end{equation}
The decoded action chunk 
\(\bar{a}_{\mathrm{final},t}= \mathcal{D}_{\mathrm a}(z^{\mathrm a}_{\mathrm{final},t})\) is executed open-loop for \(M\) control steps before the policy is queried again at the next chunk boundary. The Stage-1 action decoder \(\mathcal{D}_{\mathrm a}\) is kept frozen during this process, so all residual-corrected latents are decoded through the same mapping learned to reconstruct demonstration action chunks. We implement \(f^\theta\)'s backbone as a Transformer conditioned on observation tokens and language-conditioned base action-latent tokens; see~\autoref{appendix-sec:network-specifications}.

To obtain the visual-dynamics target, at each iteration, we run \(B\) parallel simulator environments under the current augmented policy, using the simulator only to obtain initial conditions and roll out the refined action chunks. We do not require training signals from the simulator such as rewards, success labels, or differentiable state transitions. For each environment instance \(i\), \(\mathcal{W}\) generates video latents \(\mathbf{V}_i=\{V_{i,t}\}_{t=1}^{T}\) conditioned on the initial observation and instruction \((s_{i,1},\ell_i)\), as described in Eq.~\eqref{eq:wm-denoise}. The frozen video adapter maps these latents to video-dynamics embeddings \(\mathbf{z}^{\mathrm v}_i=\mathcal{B}_{\mathrm v}(\mathbf{V}_i)\), while the augmented policy produces action-latent trajectories \(\mathbf{z}^{\mathrm a}_{\mathrm{final},i}=\{z^{\mathrm a}_{\mathrm{final},i,t}\}_{t=1}^{T}\). 

Finally, we minimize \(\mathcal{L}_{\mathrm{contrastive}}\) from~ Eq.~\eqref{eq:alignment_loss} between \(\mathbf{z}_{i}^{\mathrm v}\) and \(\mathbf{z}^{\mathrm a}_{\mathrm{final}, i}\) across the \(B\) parallel rollouts, where latents from the same rollout form positive pairs and latents from different rollouts serve as negatives, following \hyperref[sub-sec:Stage1]{Stage 1}. Gradients are taken only through \(f^\theta\), aligning the augmented policy action latents with WM-induced dynamics while keeping the VLA and Stage-1 modules frozen.

\vspace{-1.5ex}
\section{Experiments}\label{sec: experiment}
\vspace{-1ex}

We design our empirical evaluation to answer the following core questions:
\begin{itemize}[leftmargin=*, labelsep=0.5em]
\vspace{-1.5ex}
    \item \hypertarget{ref:Q1}{\textbf{Q1:}} How effectively does \textsc{World2Act} improve strong VLA baselines compared with prior post-training methods on complex manipulation tasks?

    \item \hypertarget{ref:Q2}{\textbf{Q2:}} How do latent action alignment and the residual policy architecture improve action representations toward target WM dynamics, and how do these gains affect manipulation performance?

    \item \hypertarget{ref:Q3}{\textbf{Q3:}} How does scaling post-training data and tasks affect generalization to unseen robotic skills?

    \item \hypertarget{ref:Q4}{\textbf{Q4:}} Does \textsc{World2Act} transfer effectively to real-world robot experiments?
\vspace{-1.5ex}
\end{itemize}

\vspace{-0.5ex}
\subsection{Main Results}
\vspace{-1.5ex}

We evaluate on $3$ simulated benchmarks: RoboCasa~\citep{nasiriany2024robocasa}, LIBERO~\citep{liu2023libero}, and Bridge-SIMPLER~\citep{zhou2025autoeval}, an extension of SIMPLER~\citep{li2025evaluating}. We use mean success rate (SR) as the primary metric. Following standard protocols~\citep{nasiriany2024robocasa, kim2026cosmos, shen2026videovla}, we run $50$ trials per task for each of $5$ seeds and report the mean SR.

\vspace{-0.5ex}
\noindent\textbf{Experimental Setup.}
We benchmark against recent VLA policies: $\pi_0$~\citep{black2024pi_0}, $\pi_{0.5}$~\citep{intelligence2025pi_}, UVA~\citep{li2025unified}, Rethink-VLA~\citep{wang2026rethinking}, OpenVLA~\citep{kim2024openvla}, OpenVLA-OFT~\citep{kim2025fine}, UWM~\citep{zhu2025unified}, UniVLA~\citep{bu2025univla}, Cosmos Policy~\citep{kim2026cosmos}, CoWVLA~\citep{yang2026chain}, FLARE~\citep{zheng2025flare}, and GR00T-N1.6~\citep{nvidia_gear_gr00t_n1_6_2025}. To isolate the post-training effect, we additionally compare \textsc{World2Act} to representative post-training baselines: a BC variant of \textsc{World2Act}, which replaces the contrastive objective with supervised behavior cloning on the target demonstrations~\citep{bjorck2025gr00t}, DreamGen~\citep{jang2025dreamgen}, VLA-RFT~\citep{li2025vla}, and Ctrl-World~\citep{guo2025ctrl}. For a fair comparison, we implement \textsc{World2Act} and all post-training methods on the same GR00T-N1.6 backbone. Because GR00T-N1.6 is pretrained on mixed cross-embodiment data~\citep{nvidia_physicalai_groot_x_2025}, its behavior may not align with the simulator task distributions. We therefore finetune GR00T-N1.6 on $1000$ expert trajectories for each simulator to obtain a task-aligned supervised reference, denoted as GR00T-N1.6-ft.

\begin{table*}[!ht]
\vspace{-1ex}
\centering
\scriptsize
\renewcommand{\arraystretch}{1.15}

\begin{minipage}[t]{0.32\linewidth}
\centering
\caption{\textbf{Results on RoboCasa.}}
\label{tab:robocasa_main}
\setlength{\tabcolsep}{2pt}
\resizebox{\linewidth}{!}{%
\begin{tabular}{l c}
\toprule
\textbf{Baselines} & \textbf{SR} \\
\midrule
$\pi_0$~\citep{black2024pi_0} & 0.625 \\
UVA~\citep{li2025unified} & 0.500 \\
Rethink-VLA~\citep{wang2026rethinking} & 0.547 \\
UWM~\citep{zhu2025unified} & 0.608 \\
FLARE~\citep{zheng2025flare} & 0.701 \\
GR00T-N1.6~\citep{nvidia_gear_gr00t_n1_6_2025} & 0.662 \\
Cosmos Policy~\citep{kim2026cosmos}  & 0.657 \\
\midrule
\rowcolor{grootgreen}
GR00T-N1.6-ft~\citep{nvidia_gear_gr00t_n1_6_2025} & 0.701 \\
\rowcolor{grootgreen}
GR00T-N1.6-ft+DreamGen~\citep{jang2025dreamgen} & 0.705 \\
\rowcolor{grootgreen}
GR00T-N1.6-ft+\textsc{World2Act} (Ours) & \textbf{0.726} \\
\bottomrule
\end{tabular}%
}
\end{minipage}
\hfill
\begin{minipage}[t]{0.33\linewidth}
\centering
\caption{\textbf{Results on LIBERO.} Detailed SR of Spatial, Object, Goal, and Long are in App.~\ref{appendix-sec:libero-results}.}
\label{tab:libero_results_compact}
\vspace{-0.8ex}
\setlength{\tabcolsep}{2pt}
\resizebox{\linewidth}{!}{%
\begin{tabular}{l c}
\toprule
\textbf{Baselines} & \textbf{Avg. SR} \\
\midrule
$\pi_{0.5}$~\citep{intelligence2025pi_} & 0.969 \\
OpenVLA-OFT~\citep{kim2025fine} & 0.971 \\
UniVLA~\citep{bu2025univla} & 0.952 \\
CoWVLA~\citep{yang2026chain} & 0.956 \\
Cosmos Policy~\citep{kim2026cosmos} & \textbf{0.985} \\
\midrule
\rowcolor{grootgreen}
GR00T-N1.6-ft~\citep{nvidia_gear_gr00t_n1_6_2025} & 0.970 \\
\rowcolor{grootgreen}
GR00T-N1.6-ft+DreamGen~\citep{jang2025dreamgen} & 0.921 \\
\rowcolor{grootgreen}
GR00T-N1.6-ft+\textsc{World2Act} (Ours) & \underline{0.981} \\
\bottomrule
\end{tabular}%
}
\end{minipage}
\hfill
\begin{minipage}[t]{0.32\linewidth}
\centering
\caption{\textbf{Results on Bridge-SIMPLER.} We evaluate on 7 tasks; full details in App.~\ref{appendix-sec:libero-results}.}
\label{tab:simplerenv_results_compact}
\setlength{\tabcolsep}{2pt}
\resizebox{\linewidth}{!}{%
\begin{tabular}{l c}
\toprule
\textbf{Baselines} & \textbf{SR} \\
\midrule
$\pi_0$~\citep{black2024pi_0} & 0.337 \\
OpenVLA~\citep{kim2024openvla} & 0.103 \\
OpenVLA-OFT~\citep{kim2025fine} & 0.438 \\
\midrule
\rowcolor{grootgreen}
GR00T-N1.6-ft~\citep{nvidia_gear_gr00t_n1_6_2025} & 0.576 \\
\rowcolor{grootgreen}
GR00T-N1.6-ft+DreamGen~\citep{jang2025dreamgen} &  0.583 \\
\rowcolor{grootgreen}
GR00T-N1.6-ft+\textsc{World2Act} (Ours) &  \textbf{0.590} \\
\bottomrule
\end{tabular}%
}
\end{minipage}

\vspace{-1ex}
\end{table*}

\noindent\textbf{Post-Training Data Synthesis.}
We use Cosmos-Predict2~\citep{agarwal2025cosmos} to synthesize approximately $1000$ imagined trajectories per simulator across RoboCasa, LIBERO, and Bridge-SIMPLER for all post-training methods, totaling roughly $3000$ synthetic trajectories. Post-training initial scenes are disjoint from evaluation scenes to prevent test leakage. Additional details are provided in Appendix~\ref{appendix:post-training}.

\noindent\textbf{Results on RoboCasa.}
\autoref{tab:robocasa_main} shows that \textsc{World2Act} achieves the best RoboCasa result, improving the same backbone GR00T-N1.6-ft from $70.1\%$ to $72.6\%$ SR, a $+2.5\%$ absolute gain. It also outperforms DreamGen under the same finetuned starting point by $+2.1\%$, showing that our imagined-trajectory post-training objective is more effective than prior synthetic-data augmentation. Against the broader set of baselines, \textsc{World2Act} leads vanilla GR00T-N1.6 by $+6.4\%$ and the WM-capable Cosmos Policy by $+6.9\%$, establishing the strongest RoboCasa performance overall.

\noindent\textbf{Results on LIBERO.}
\autoref{tab:libero_results_compact} reports average SR on LIBERO. \textsc{World2Act} achieves a $98.1\%$ SR, while DreamGen degrades GR00T-N1.6-ft to $92.1\%$, suggesting IDM noise from WM artifacts (WM rollouts raise IDM action MSE by $18\%$; \autoref{tab:visual-artifacts}, Appendix~\ref{sec:idm_artifacts}). Our method therefore yields a $+6.0\%$ absolute gain over DreamGen on the same backbone. It also surpasses $\pi_{0.5}$, OpenVLA-OFT, CoWVLA, and UniVLA, and comes within $0.4\%$ of the best reported Cosmos Policy result.

\noindent\textbf{Results on Bridge-SIMPLER.}
\autoref{tab:simplerenv_results_compact} reports average success on Bridge-SIMPLER. We use AutoEval Bridge-SIMPLER~\citep{zhou2025autoeval} because it extends the original WidowX-based SIMPLER Bridge setup with three additional tasks, giving a broader evaluation for the WidowX embodiment. Consistent with RoboCasa and LIBERO, \textsc{World2Act} improves GR00T-N1.6-ft from $57.6\%$ to $59.0\%$ SR, a $+1.4\%$ absolute gain, and also outperforms DreamGen ($58.3\%$), achieving the best overall result.

\begin{figure}[!ht]
    \centering
    \vspace{-2.3ex}
    \begin{subfigure}{0.50\linewidth}
        \centering
        \includegraphics[width=\linewidth]{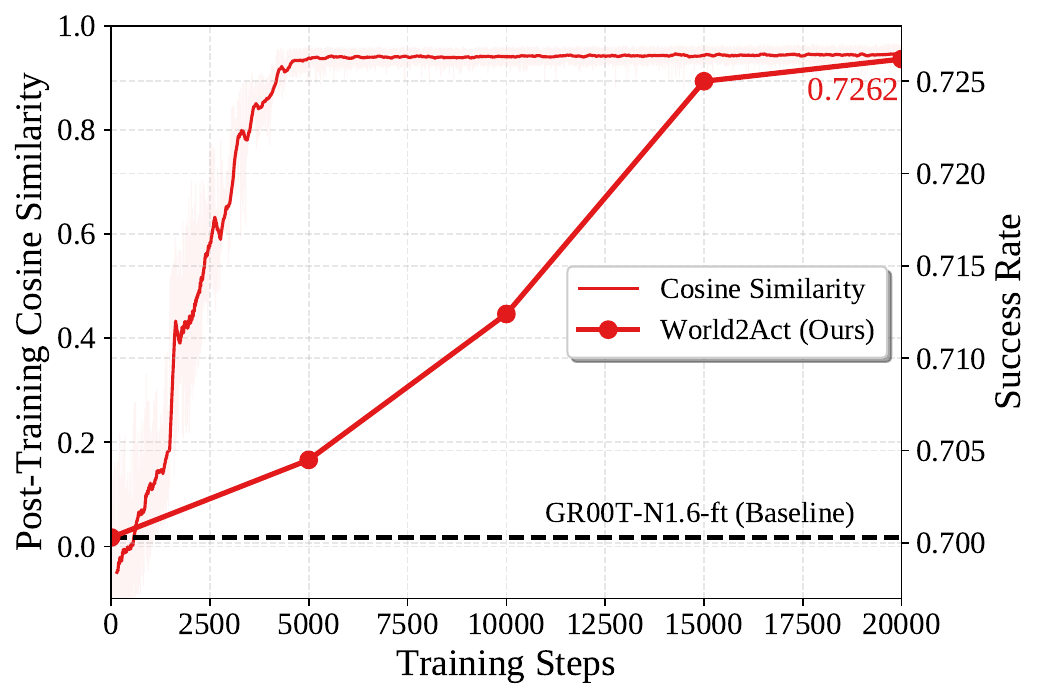}
        \vspace{-4ex}
        \caption{\textbf{Correlation between post-training cosine similarity and success rate.} Higher video--action alignment yields better manipulation performance.}
        \label{fig:cosine-corre}
    \end{subfigure}
    \hfill
    \begin{subfigure}{0.49\linewidth}
        \centering
        \includegraphics[width=\linewidth]{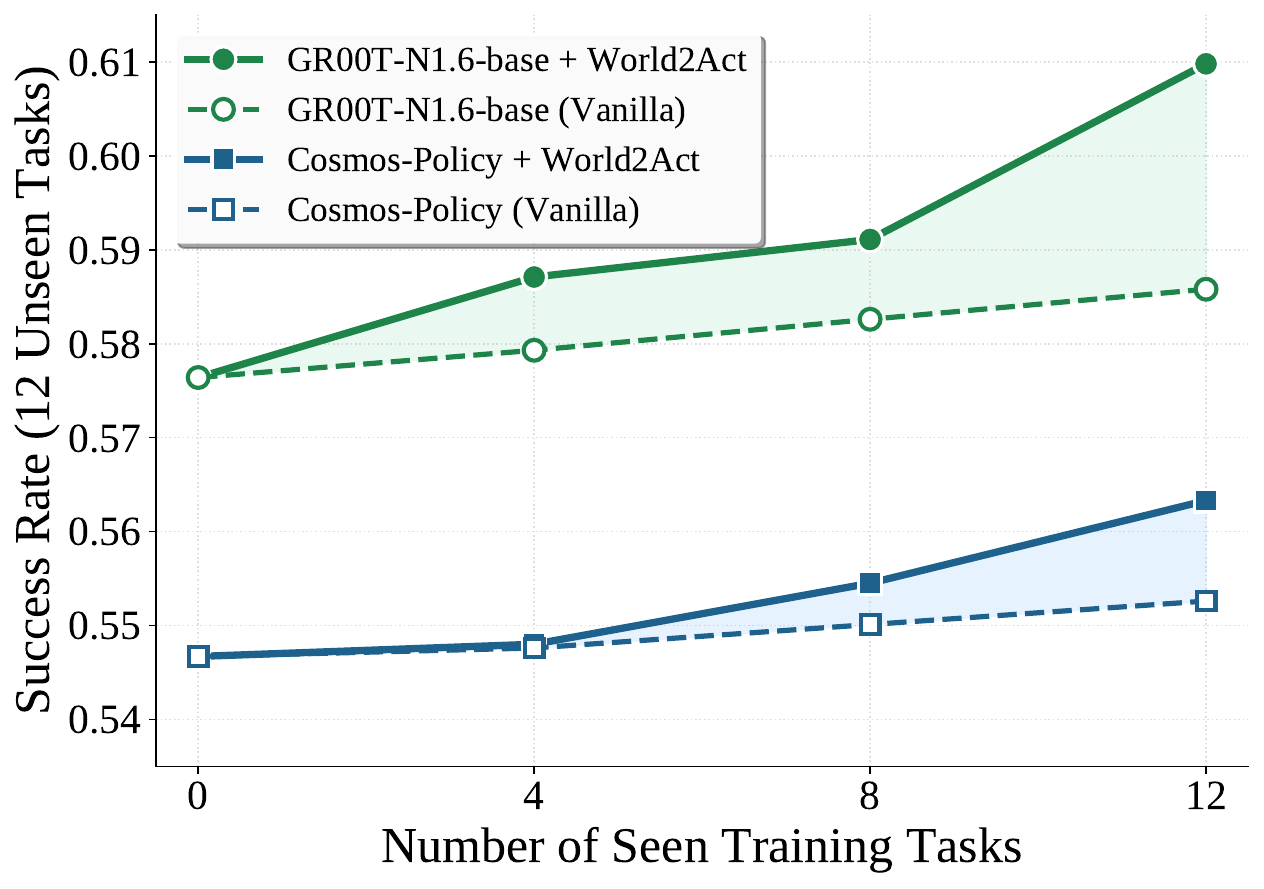}
        \vspace{-4ex}
        \caption{\textbf{Cross-task generalization.} Scaling \textsc{World2Act} on more seen tasks improves performance when evaluated on the remaining unseen tasks.}
        \label{fig:generalization_seen_tasks}
    \end{subfigure}
    \vspace{-2.7ex}
    \caption{\textbf{Post-training scaling and generalization.} \textit{Left:} Cosine similarity correlates with downstream success rate. \textit{Right:} Unseen-task performance as seen-task diversity increases.}
    \label{fig:main_scaling_figure}
    \vspace{-1.5ex}
\end{figure}

\noindent \textbf{Cross-Task Generalization.}
We split $24$ RoboCasa tasks into $12$ seen/$12$ unseen tasks and scale post-training from $0$ to $12$ seen tasks (details in Appendix~\ref{appendix-subsec:cross-task-generalization}). To isolate cross-task generalization, we evaluate base VLAs without task-specific finetuning. On the unseen split, \textsc{World2Act} consistently improves with task diversity, yielding $+2.4\%$ on GR00T-N1.6-base and $+1.1\%$ on Cosmos Policy (Fig.~\ref{fig:generalization_seen_tasks}). Since the vanilla VLAs never observe unseen-task trajectories, these gains suggest that generalized WM dynamics effectively transfer to novel skills.

\begin{table}[!ht]
\vspace{-2.6ex}
\centering
\footnotesize
\setlength{\tabcolsep}{3pt}
\renewcommand{\arraystretch}{0.92}

\begin{minipage}[t]{0.38\linewidth}
\centering
\caption{\textbf{Post-training comparison.}
All rows use GR00T-N1.6-ft as the base model; `+' denotes post-training.}
\label{tab:post_training_compact}
\begin{tabularx}{\linewidth}{@{}>{\raggedright\arraybackslash}Xc@{}}
\toprule
\textbf{Method} & \textbf{SR} \\
\midrule
GR00T-N1.6-ft~\citep{nvidia_gear_gr00t_n1_6_2025} & 0.701 \\
+\textsc{World2Act} w/ BC~\citep{bjorck2025gr00t} & 0.704 \\
+DreamGen~\citep{jang2025dreamgen} & 0.705 \\
+VLA-RFT~\citep{li2025vla} & 0.710 \\
+Ctrl-World~\citep{guo2025ctrl} & 0.698 \\
\midrule
\rowcolor{grootgreen}
+\textsc{World2Act} (Ours) & \textbf{0.726} \\
\bottomrule
\end{tabularx}
\end{minipage}
\hfill
\begin{minipage}[t]{0.58\linewidth}
\centering
\caption{\textbf{Adapter and latent objective ablation studies.}
Latent obj. denotes the contrastive objective and similarity design used to align WM video latents with action latents.}
\label{tab:wm_adapter_compact}
\begin{tabularx}{\linewidth}{@{}l>{\raggedright\arraybackslash}Xcc@{}}
\toprule
\textbf{Study} & \textbf{Variant} & \textbf{Time} & \textbf{SR} \\
\midrule
Latent obj. 
& Single InfoNCE~\citep{lee2025class} + chunk sim. 
& -- & 0.720 \\
& Marginal~\citep{saunshi2019theoretical} + chunk sim. 
& -- & 0.707 \\
& Bi InfoNCE~\citep{oord2018representation} + global sim. 
& -- & 0.693 \\
\midrule
Adapter 
& LoRA, $r{=}16$ 
& 14.6h & 0.714 \\
& LoRA, $r{=}32$ 
& 15.3h & 0.721 \\

\midrule
\rowcolor{grootgreen}
Ours 
& Residual + Bi InfoNCE + chunk sim. 
& \textbf{6.8h} & \textbf{0.726} \\
\bottomrule
\end{tabularx}
\end{minipage}

\vspace{-2.4ex}
\end{table}

\noindent\textbf{Comparison of Post-Training Methods.} \autoref{tab:post_training_compact} compares post-training methods using GR00T-N1.6-ft as the base model. \textsc{World2Act} improves RoboCasa success rate from $70.1\%$ to $72.6\%$, outperforming BC on expert trajectories ($70.4\%$) and prior post-training baselines, including DreamGen ($70.5\%$), VLA-RFT ($71.0\%$), and Ctrl-World ($69.8\%$). These results suggest that latent-space post-training provides a more effective adaptation signal than existing alternatives.

\vspace{-1.5ex}
\subsection{Ablation Study}
\vspace{-1ex}

\noindent\textbf{Correlation Between Cosine Similarity and Success Rate.} Fig.~\ref{fig:cosine-corre} shows that video--action cosine similarity ($\text{sim}(\mathbf{z}_i^\text{v}, \mathbf{z}_i^\text{a})$ from Eq.~\eqref{eq:alignment_loss}) increases alongside downstream SR during post-training. For \textsc{World2Act}, cosine similarity rapidly climbs and stabilizes around $0.95$ after 5K steps, while SR rises from $70.1\%$ to $72.6\%$. This trend suggests that post-training strengthens video--action alignment and that stronger alignment is associated with improved manipulation performance.

\noindent\textbf{Scaling Post-Training Trajectories.}
Fig.~\ref{fig:scaling-trajectories} plots SR on RoboCasa as the number of post-training trajectories $N$ increases on the same GR00T-N1.6-ft base model: red denotes \textsc{World2Act} latent post-training, blue denotes DreamGen pixel-space post-training, and the gray dotted line marks the GR00T-N1.6-ft baseline.
\textsc{World2Act} scales monotonically from $70.1\%$ to $72.6\%$ as $N$ increases from $0$ to $1000$, while DreamGen is non-monotonic, dropping to $69.1\%$ at $N{=}500$ before recovering to $70.5\%$, suggesting that pixel-space pseudo-labeling is more sensitive to post-training noise.


\noindent\textbf{Residual Policy vs. LoRA Finetuning.}~\autoref{tab:wm_adapter_compact} compares our residual policy against LoRA finetuning on performance and speed. Compared with LoRA, our residual policy achieves higher SR at much lower cost: $72.6\%$ vs. $72.1\%$ for LoRA $r{=}32$, with $2.25{\times}$ faster training ($6.8$h vs. $15.3$h). This supports freezing the VLA and learning only a lightweight action-space correction.

\noindent\textbf{Contrastive Loss Ablation Study.}
The latent-objective ablation in~\autoref{tab:wm_adapter_compact} shows that both bidirectional contrast and chunk-level temporal alignment are important. Replacing our bidirectional InfoNCE with unidirectional InfoNCE~\citep{lee2025class} drops SR from $72.6\%$ to $72.0\%$, while using a marginal loss~\citep{saunshi2019theoretical} further reduces it to $70.7\%$. Collapsing the temporally aligned chunk-wise similarity into a single global trajectory similarity performs worst at $69.3\%$, indicating that fine-grained video--action alignment, rather than coarse trajectory-level matching, is critical for manipulation.

\begin{figure}[!ht]
    \centering
    \vspace{-1ex}
    \begin{minipage}[t]{0.65\textwidth}
        \vspace{0pt}
        \centering
        \includegraphics[
            width=\linewidth,
            trim={0pt 0pt 0pt 5pt},
            clip
        ]{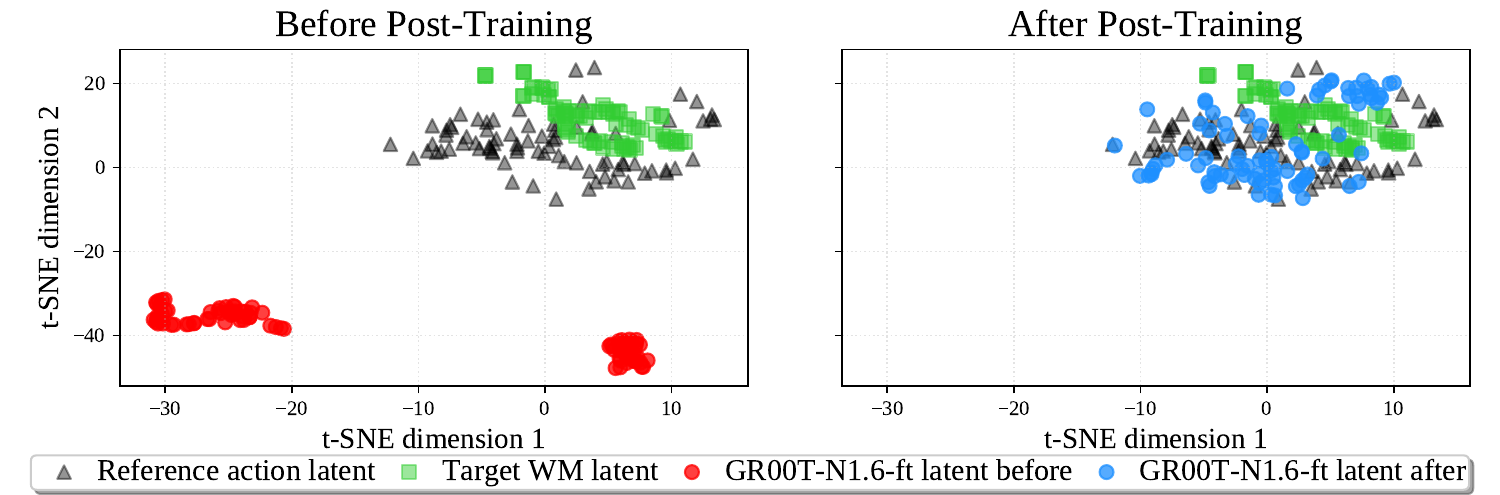}
        \caption{
            \textbf{Action-latent t-SNE visualization.}
        }
        \label{fig:opendrawer_tsne}
    \end{minipage}
    \hfill
    \begin{minipage}[t]{0.34\textwidth}
        \vspace{0pt}
        \centering
        \captionof{table}{
            \textbf{Action-space error.}
            MSE is computed between predicted and reference-expert actions over 3 tasks; lower is better.
        }
        \label{tab:action_mse}

        \small
        \begin{tabular}{@{}lc@{}}
            \toprule
            \textbf{Method} & \textbf{MSE} $\downarrow$ \\
            \midrule
            GR00T-N1.6-ft
                & 0.034 \\
            +\textsc{World2Act} (Ours)
                & \textbf{0.021} \\
            \bottomrule
        \end{tabular}
    \end{minipage}
    \vspace{-2ex}
\end{figure}

\noindent\textbf{Action Validity Analysis.}
We assess how Stage-2 \textsc{World2Act} latent alignment affects action-manifold compatibility. Fig.~\ref{fig:opendrawer_tsne} shows, on \textit{OpenDrawer}, that predicted-action latents after post-training overlap more with reference-action and target-WM latent regions than before. Table~\ref{tab:action_mse} further reports action-space MSE between predicted actions and temporally matched held-out expert actions, averaged over \textit{OpenDrawer}, \textit{PnPCounterToCab}, and \textit{TurnOnStove}; the error drops from $0.034$ to $0.021$, a $0.013$ decrease ($38.2\%$ relative). These results are consistent with our weak-supervision view: Stage-2 residual updates move action latents toward WM dynamics priors while keeping decoded actions near the expert action manifold. Additional analyses are in Appendix~\ref{app:extra_latent_visualization}.

\begin{figure}[!ht]
    \vspace{-1.2ex}
    \centering
    \begin{minipage}{0.46\textwidth}
        \centering
        \vspace{0.5ex}\includegraphics[width=0.96\linewidth]{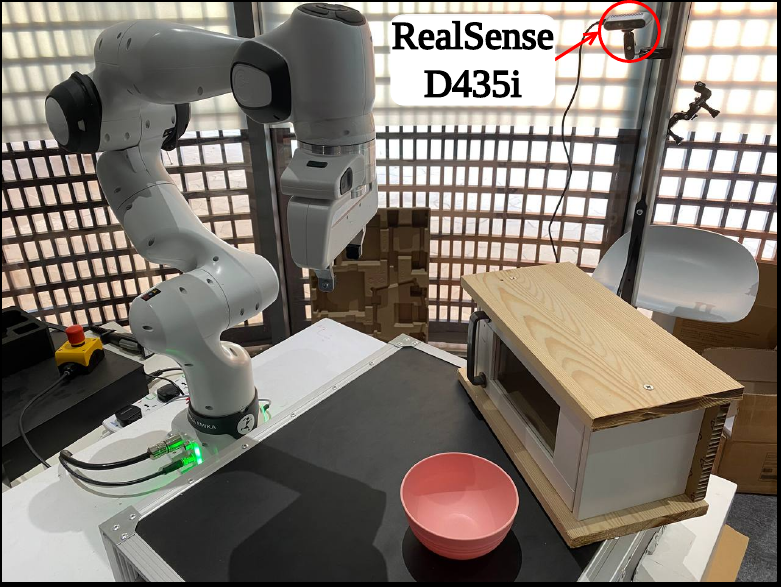}
        \vspace{-0.4ex}
        \caption{\textbf{Robot Setup.}}
        \vspace{-2.8ex}\label{fig:robot_setup}
    \end{minipage}
    \hfill
    \begin{minipage}{0.53\textwidth}
        \centering
        \includegraphics[width=\linewidth, trim={30pt 30pt 30pt 30pt}, clip]{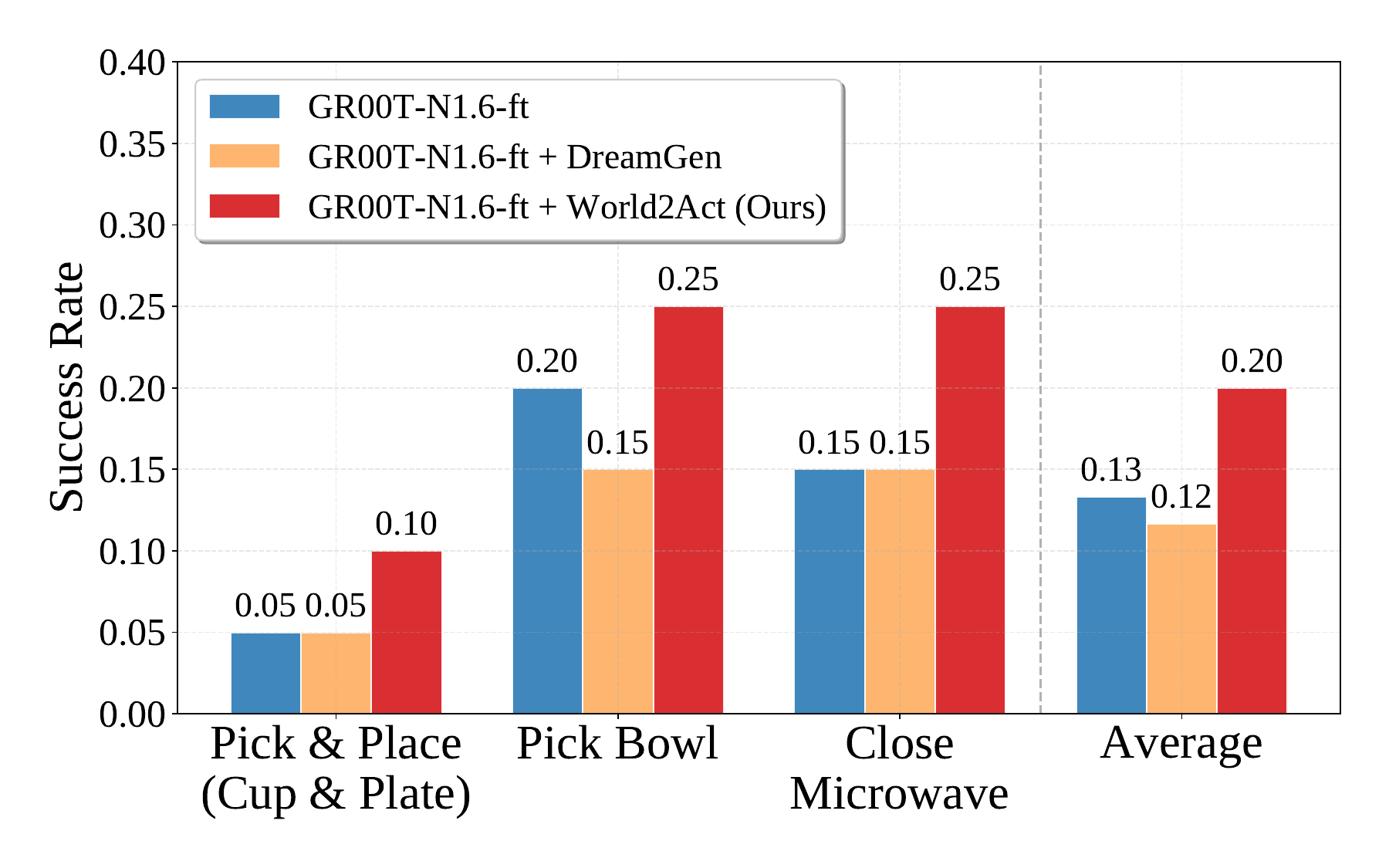}
        \vspace{-3ex}\caption{\textbf{Real-world robot success rates.}}
        \vspace{-2.3ex}\label{fig:success_rates}
    \end{minipage}
\end{figure}

\subsection{Real Robot Experiments}\label{subsec: real-robot}
\vspace{-1.5ex}

\noindent\textbf{Setup.} We evaluate \textsc{World2Act} on a Franka Research 3 across three real-world tasks: \textit{Pick cup and place on plate}, \textit{Pick up bowl}, and \textit{Close microwave}. We start from GR00T-N1.6-ft with $20$ real demonstrations per task, collected using GELLO~\citep{wu2024gello}, and further train with $100$ WM-generated trajectories. Observations come from an external camera (Fig.~\ref{fig:robot_setup}), and we run $20$ trials per task.

\noindent\textbf{Results.} Fig.~\ref{fig:success_rates} summarizes real-world evaluations. \textsc{World2Act} improves GR00T-N1.6-ft by $6.7\%$ on average and outperforms DreamGen by $8.3\%$, suggesting that latent WM-guided post-training transfers more effectively to physical robots than synthetic-data augmentation.

\begin{figure}[!ht]
    \vspace{-2ex}
    \centering
    \raggedright 
    \includegraphics[width=\linewidth]{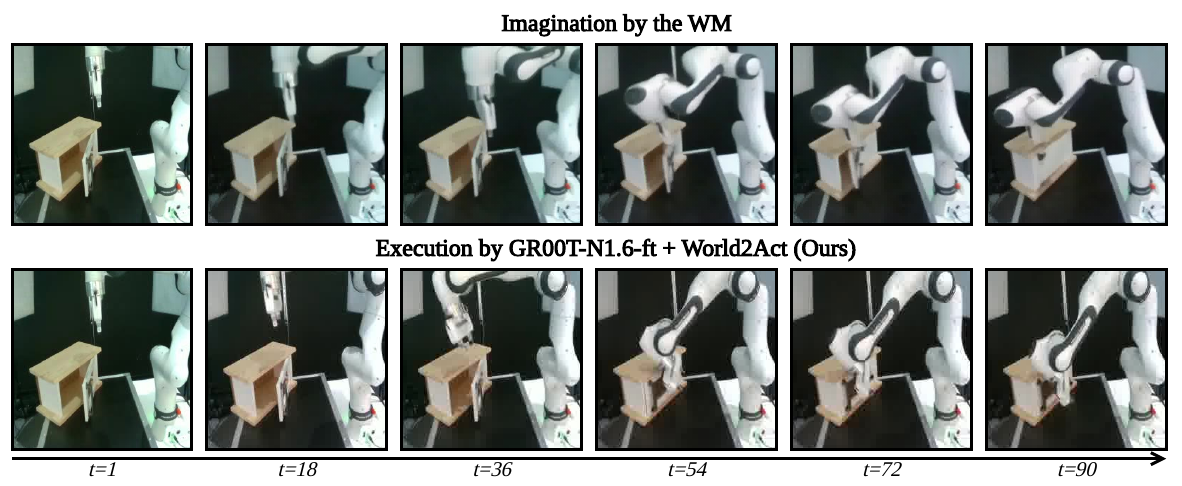}
    \vspace{-3.4ex}
    \caption{\textbf{Qualitative results of our WM-to-VLA framework.} \textit{Top:} A trajectory generated by the WM. \textit{Bottom:} The corresponding real-robot execution of the same task utilizing GR00T-N1.6-ft + \textsc{World2Act}. $t$ denotes the timestep for each frame.}
    \label{fig:wm_qualitative}
    \vspace{-1.3ex}
\end{figure}

\noindent\textbf{Qualitative Evaluation.} Fig.~\ref{fig:wm_qualitative} compares a representative WM rollout with the corresponding robot execution for the \textit{Close microwave} task. Both the imagined rollout and the physical execution successfully achieve the task goal. Notably, the final frame of the imagined sequence ($t{=}90$) exhibits visual artifacts, such as the handle becoming invisible. However, because our design utilizes the latent dynamics of the WM, it remains robust to these imperfect rollouts, as evidenced by the successful trial.  Additional robot setup details and demonstrations are provided in Appendix~\ref{appendix-subsec:real-world-experiments}.

\noindent\textbf{Remarks.}
\textsc{World2Act} distills latent WM dynamics into a GR00T-N1.6 policy~(Fig.~\ref{fig:cosine-corre}), improving simulation success across RoboCasa, LIBERO, and Bridge-SIMPLER over strong VLA and post-training baselines~(Tables~\ref{tab:robocasa_main},~\ref{tab:libero_results_compact},~\ref{tab:simplerenv_results_compact},~\ref{tab:post_training_compact}) while exposing the fragility of pixel-space WM supervision: DreamGen degrades GR00T-N1.6-ft on LIBERO, suggesting pseudo-actions from decoded WM rollouts can be corrupted by visual artifacts~(see failure-case study in Appendix~\ref{appendix-sec:failure-case})~(\hyperlink{ref:Q1}{\textbf{Q1}}). Furthermore, our residual policy and contrastive alignment are important: they increase the overlap between predicted-action latents and the reference-action/target-WM latent regions~(Fig.~\ref{fig:opendrawer_tsne}) while reducing action-space error~(Table~\ref{tab:action_mse}), leading to higher downstream success rates~(\autoref{tab:wm_adapter_compact})~(\hyperlink{ref:Q2}{\textbf{Q2}}). Our post-training paradigm scales favorably, showing consistent gains with increasing trajectory volume~(Fig.~\ref{fig:scaling-trajectories}) and strong cross-task generalization~(Fig.~\ref{fig:generalization_seen_tasks})~(\hyperlink{ref:Q3}{\textbf{Q3}}). Finally, real-world experiments show that our post-training method transfers effectively to physical robot deployments, improving success rates while remaining stable to imperfect synthetic rollouts~(Figs.~\ref{fig:success_rates},~\ref{fig:wm_qualitative})~(\hyperlink{ref:Q4}{\textbf{Q4}}).
\vspace{-2ex}
\section{Discussion}\label{sec: discussion}
\vspace{-1.5ex}
We introduced \textsc{World2Act}, a latent post-training framework that transfers generalizable dynamics priors from WMs to a GR00T-N1.6 policy, avoiding fragile pixel-level supervision. Experiments across benchmarks show consistent performance improvements over pixel-based post-training and other strong baselines, highlighting the promise of using WM latents as reusable dynamics priors.

\noindent\textbf{Limitations.} \textsc{World2Act} has limitations. First, contact-rich, non-prehensile tasks~\cite{shirai2026learning} exhibit complex contact-driven dynamics that remain difficult for WMs to capture: small errors in object motion or contact evolution can compound in WM rollouts, making the resulting priors less transferable. Second, real-world success remains modest, suggesting a persistent domain gap where complex physical dynamics are insufficiently captured by current WMs and VLAs~\citep{motamed2026generative, yu2026dm0}. Third, our chunk-level video--action coupling uses temporal correspondence between imagined and executed trajectories, motivating future work relaxing this alignment for more flexible execution dynamics.


\bibliography{main}

\newpage
\appendix

\appendix

\noindent This appendix provides additional details and extended evaluations to support the main paper. The material is organized as follows:
\begin{itemize}
    \item \textbf{\autoref{appendix-sec:libero-results}} reports additional results on the simulation benchmarks in the main paper.
    \item \textbf{\autoref{appendix-sec:experimental-details}} provides experimental details for \textsc{World2Act}, including Stage 1 pretraining, cross-task evaluation, real-world deployment, and inference-speed analysis.
    \item \textbf{\autoref{appendix-sec:network-specifications}} specifies the network architectures and hyperparameters for our proposed policy components.
    \item \textbf{\autoref{appendix-sec:data-wm}} describes the data construction pipeline for world-model training, including atomic-skill segmentation, LLM-based synchronization, dataset statistics, and multi-view stitching.
    \item \textbf{\autoref{appendix-sec:wm-evaluation}} details the training configurations and the evaluation of the video diffusion world models used in our framework.
    \item \textbf{\autoref{appendix-sec:failure-case}} analyzes representative failure cases, including world-model hallucinations and transfer failures.
    \item \textbf{\autoref{appendix-sec:broader-impact}} and \textbf{\autoref{appendix-sec:extra-related-work}} discuss the broader impact of our work and provide further clarifications on related literature, respectively.
\end{itemize}

\section{Supplementary Experiment Results}
\label{appendix-sec:libero-results}

In this appendix section, we provide additional experimental results on LIBERO, RoboCasa, and Bridge-SIMPLER.  Beyond the main backbone studied in the paper, we evaluate \textsc{World2Act} on Cosmos Policy, a strong VLA with built-in world-modeling capability. This experiment tests whether \textsc{World2Act} can still improve policies that already incorporate WM structure.

\begin{table}[!ht]
\caption{\textbf{Detailed LIBERO results.}}
\label{tab:libero_main_results}
\centering
\small
\setlength{\tabcolsep}{10pt}
\renewcommand{\arraystretch}{1.15}
\resizebox{\linewidth}{!}{%
\begin{tabular}{l c c c c c}
\toprule
\textbf{Baselines} & \textbf{Spatial} & \textbf{Object} & \textbf{Goal} & \textbf{Long} & \textbf{Average} \\
\midrule
$\pi_0$~\citep{black2024pi_0} & 0.968 & 0.988 & 0.958 & 0.852 & 0.942 \\
$\pi_{0.5}$~\citep{intelligence2025pi_} & 0.988 & 0.982 & 0.980 & 0.924 & 0.969 \\
OpenVLA-OFT~\citep{kim2025fine} & 0.976 & 0.984 & 0.979 & 0.945 & 0.971 \\
UniVLA~\citep{bu2025univla} & 0.965 & 0.968 & 0.956 & 0.920 & 0.952 \\
CoWVLA~\citep{yang2026chain} & 0.972 & 0.978 & 0.946 & 0.928 & 0.956 \\
\midrule
\rowcolor{cosmosblue}
Cosmos Policy~\citep{kim2026cosmos} & 0.981 & 1.000 & 0.982 & 0.976 & 0.985 \\
\rowcolor{cosmosblue}
Cosmos Policy~\citep{kim2026cosmos} + \textsc{World2Act} (Ours) & 0.980 & 1.000 & 0.983 & 0.980 & \textbf{0.986} \\
\midrule
\rowcolor{grootgreen}
GR00T-N1.6-ft~\citep{nvidia_gear_gr00t_n1_6_2025} & 0.977 & 0.984 & 0.975 & 0.943 & 0.970 \\
\rowcolor{grootgreen}
GR00T-N1.6-ft~\citep{nvidia_gear_gr00t_n1_6_2025} + DreamGen~\citep{jang2025dreamgen} & 0.991 & 0.826 & 0.992 & 0.876 & 0.921 \\
\rowcolor{grootgreen}
GR00T-N1.6-ft~\citep{nvidia_gear_gr00t_n1_6_2025} + \textsc{World2Act} (Ours) & 0.995 & 1.000 & 0.988 & 0.940 & \textbf{0.981} \\
\bottomrule
\end{tabular}
}
\end{table}

\noindent\textbf{Performance on LIBERO.}
\autoref{tab:libero_main_results} summarizes results across the spatial, object, goal, and long-horizon LIBERO suites. On Cosmos Policy, \textsc{World2Act} improves the average success rate from 98.5\% to 98.6\%. Although the absolute gain is modest, Cosmos Policy already achieves near-perfect performance on LIBERO, a near-saturated benchmark, making further improvements difficult. The positive gain therefore suggests that \textsc{World2Act} can further improve performance even when the base policy already benefits from world-modeling capability.

\begin{table}[!ht]
\caption{\textbf{Extra RoboCasa results.}}
\label{tab:robocasa_main_appendix}
\centering
\small
\setlength{\tabcolsep}{10pt}
\renewcommand{\arraystretch}{1.15}
\begin{tabular}{l c}
\toprule
\textbf{Baselines} & \textbf{SR} \\
\midrule
\rowcolor{cosmosblue}
Cosmos Policy~\citep{kim2026cosmos} & 0.657 \\
\rowcolor{cosmosblue}
Cosmos Policy~\citep{kim2026cosmos} + \textsc{World2Act} (Ours) & \textbf{0.663} \\
\midrule
\rowcolor{grootgreen}
GR00T-N1.6-ft~\citep{nvidia_gear_gr00t_n1_6_2025} & 0.701 \\
\rowcolor{grootgreen}
GR00T-N1.6-ft~\citep{nvidia_gear_gr00t_n1_6_2025} + DreamGen~\citep{jang2025dreamgen} & 0.705 \\
\rowcolor{grootgreen}
GR00T-N1.6-ft~\citep{nvidia_gear_gr00t_n1_6_2025} + \textsc{World2Act} (Ours) & \textbf{0.726} \\
\bottomrule
\end{tabular}
\end{table}

\noindent\textbf{Performance on RoboCasa.}
\autoref{tab:robocasa_main_appendix} reports the corresponding results on RoboCasa.
Consistent with the LIBERO results, \textsc{World2Act} improves Cosmos Policy from 65.7\% to 66.3\%.
Again, the gain is moderate, but this is precisely the regime of interest: Cosmos Policy already has WM capability, yet post-training with \textsc{World2Act} still increases downstream task success.
This result supports the view that \textsc{World2Act} is complementary to existing WM-capable VLAs rather than only being useful for policies without an explicit world-modeling component.

Overall, the supplementary results across LIBERO and RoboCasa reinforce our main claim: using a world model to post-train a VLA can continue to improve policy performance, even when the underlying VLA already has WM capability.

\begin{table}[!ht] \caption{\textbf{Detailed Bridge-SIMPLER results.}} \label{tab:bridge_simpler_detailed_results} \centering \small \setlength{\tabcolsep}{4pt} \renewcommand{\arraystretch}{1.15} \resizebox{\linewidth}{!}{
\begin{tabular}{l c c c c c c c c} \toprule \textbf{Baselines} & \shortstack{\textbf{Put Spoon}\\\textbf{on Towel}} & \shortstack{\textbf{Stack Green}\\\textbf{on Yellow}} & \shortstack{\textbf{Put Carrot}\\\textbf{on Plate}} & \shortstack{\textbf{Put Eggplant}\\\textbf{in Basket}} & \shortstack{\textbf{Open}\\\textbf{Drawer}} & \shortstack{\textbf{Close}\\\textbf{Drawer}} & \shortstack{\textbf{Put Eggplant}\\\textbf{in Sink}} & \textbf{Average} 
\\ \midrule $\pi_0$~\citep{black2024pi_0} & 0.292 & 0.000 & 0.166 & 0.624 & 0.676 & 0.480 & 0.124 & 0.337 \\ 
OpenVLA~\citep{kim2024openvla} & 0.000 & 0.000 & 0.000 & 0.042 & 0.644 & 0.036 & 0.000 & 0.103 \\ 
OpenVLA-OFT~\citep{kim2025fine} & 0.342 & 0.308 & 0.292 & 0.726 & 0.338 & 0.742 & 0.320 & 0.438 \\
\midrule \rowcolor{grootgreen} GR00T-N1.6-ft~\citep{nvidia_gear_gr00t_n1_6_2025} & 0.734 & 0.044 & 0.624 & 0.372 & 0.916 & 0.984 & 0.358 & 0.576 \\ \rowcolor{grootgreen} 
GR00T-N1.6-ft~\citep{nvidia_gear_gr00t_n1_6_2025} + DreamGen~\citep{jang2025dreamgen} & 0.718 & 0.048 & 0.646 & 0.424 & 0.928 & 0.976 & 0.342 & 0.583 \\
\rowcolor{grootgreen} GR00T-N1.6-ft~\citep{nvidia_gear_gr00t_n1_6_2025} + \textsc{World2Act} (Ours) & 0.768 & 0.042 & 0.664 & 0.386 & 0.916 & 0.990 & 0.362 & \textbf{0.590} \\ \bottomrule \end{tabular} } 
\end{table}

\noindent\textbf{Detailed Bridge-SIMPLER Results.}
\autoref{tab:bridge_simpler_detailed_results} reports the per-task breakdown on the AutoEval Bridge-SIMPLER~\citep{zhou2025autoeval} suite, which extends the original WidowX-based SIMPLER Bridge setup with additional drawer-manipulation tasks and a sink-placement task. This table complements the LIBERO and RoboCasa supplementary results by focusing on the WidowX-based Bridge-SIMPLER evaluation rather than an additional Cosmos Policy backbone study. \textsc{World2Act} achieves the best average SR, improving GR00T-N1.6-ft from $57.6\%$ to $59.0\%$ and outperforming DreamGen ($58.3\%$).

\section{Experimental Details}
\label{appendix-sec:experimental-details}

In this section, we provide implementation details that support the empirical evaluations in the main paper. Specifically, we describe Stage 1 pretraining, cross-task evaluation, real-world deployment, and inference-speed analysis for \textsc{World2Act}.

\subsection{Stage 1 Pretraining Details}
\label{appendix-subsec:stage1-pretraining}

To train the action and video adapters $\mathcal{B}_{\mathrm a}$ and $\mathcal{B}_{\mathrm v}$, we start with a frozen world model $\mathcal{W}$ that has been finetuned on our processed datasets (see~\autoref{appendix-sec:data-wm}). We use synchronized expert demonstrations consisting of the video $\mathbf{Q}_i$, instruction $\ell_i$, and ground-truth action sequence $\mathbf{a}_{i,\mathrm{gt}}$. Each demonstration video is encoded by the frozen WM video-VAE encoder to obtain the corresponding latent visual-dynamics trajectory:
\begin{equation}
    \mathbf{V}^{\mathrm{gt}}_i
    =
    \mathcal{E}_{\mathrm{VAE}}(\mathbf{Q}_i),
\end{equation}
where $\mathbf{V}^{\mathrm{gt}}_i=\{V^{\mathrm{gt}}_{i,t}\}_{t=1}^{T}$ lies in the same structured video-VAE latent space used by the world model. These encoded video latents serve as the video-side anchors for Stage 1 alignment. Since $\mathbf{Q}_i$ and $\mathbf{a}_{i,\mathrm{gt}}$ come from the same demonstration, each video-latent chunk is temporally aligned with the corresponding expert action chunk. 

We map the video-latent chunks into the shared latent space using the video adapter, \(z^{\mathrm v}_{i,t} =\mathcal{B}_{\mathrm v}(V^{\mathrm{gt}}_{i,t}),\) and map the corresponding ground-truth action chunks $\bar{a}^{\mathrm{gt}}_{i,t}$ into the same space using the action adapter, \(z^{\mathrm a}_{i,t}= \mathcal{B}_{\mathrm a}(\bar{a}^{\mathrm{gt}}_{i,t}).\) Stage 1 training optimizes the action reconstruction loss $\mathcal{L}_{\text{recon.}}$ and the video--action contrastive loss $\mathcal{L}_{\text{contrastive}}$, as described in~\autoref{sub-sec:Stage1}, for 30K steps. We do not decode $\mathbf{V}^{\mathrm{gt}}_i$ or use pixel-space video reconstruction as a training signal.

During minibatch construction, we use a batch size $B=16$ with a hard-negative ratio of $0.25$, where hard negatives are drawn from different demonstrations of the same task and the remaining negatives are drawn from different tasks.

\begin{figure}[!ht]
    \centering
    \includegraphics[width=\linewidth, trim={20pt 20pt 10pt 30pt}, clip]{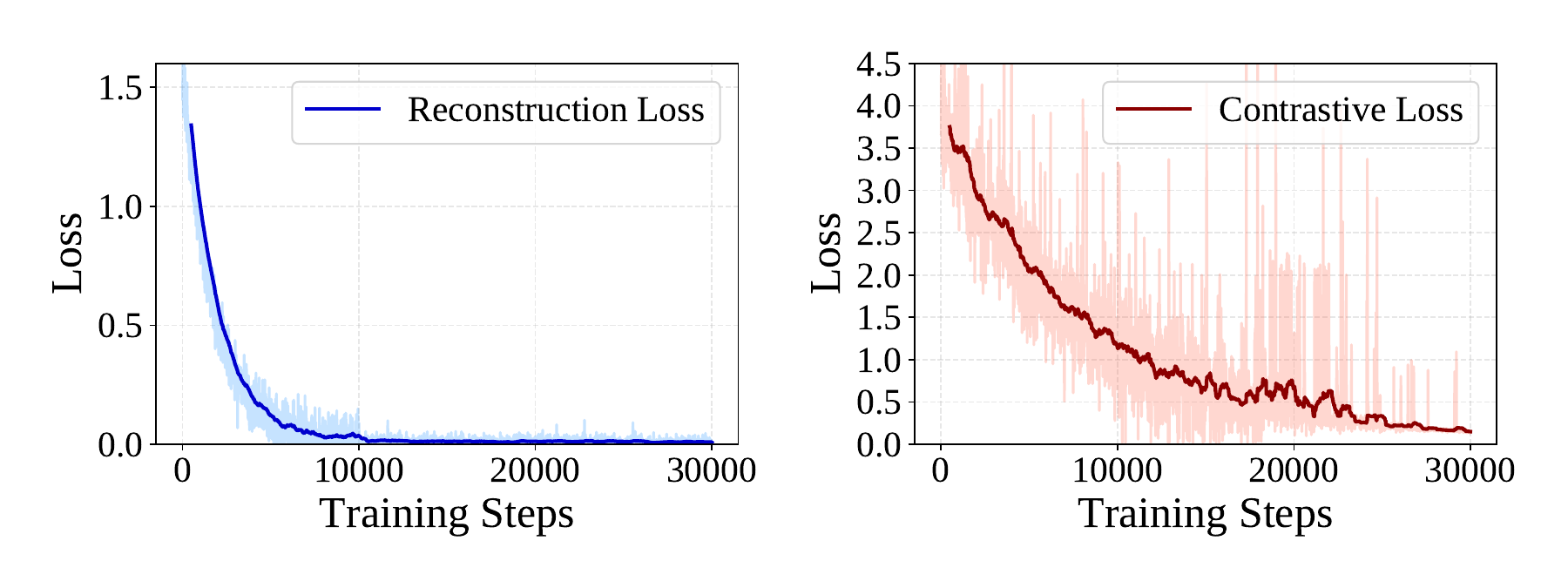}
    \caption{\textbf{Stage 1 pretraining losses.} Training curves for $\mathcal{L}_{\text{recon.}}$ and $\mathcal{L}_{\text{contrastive}}$ (defined in~\autoref{sub-sec:Stage1} of our main paper) over 30K steps. Both losses converge stably, demonstrating effective cross-modal alignment between the video and action representations.}
    \label{fig:stage1_loss}
\end{figure}

As illustrated in~\autoref{fig:stage1_loss}, both losses converge stably under our configuration: $\mathcal{L}_{\text{recon.}}$ stabilizes at approximately 0.01, while $\mathcal{L}_{\text{contrastive}}$ plateaus near 0.10. This provides optimization evidence that the action latents preserve low-level kinematic structure while the video and action modalities become well-aligned in the shared latent space.

\subsection{Post-Training Data Synthesis}
\label{appendix:post-training}

\begin{figure}[!ht]
    \centering
    \includegraphics[width=\linewidth]{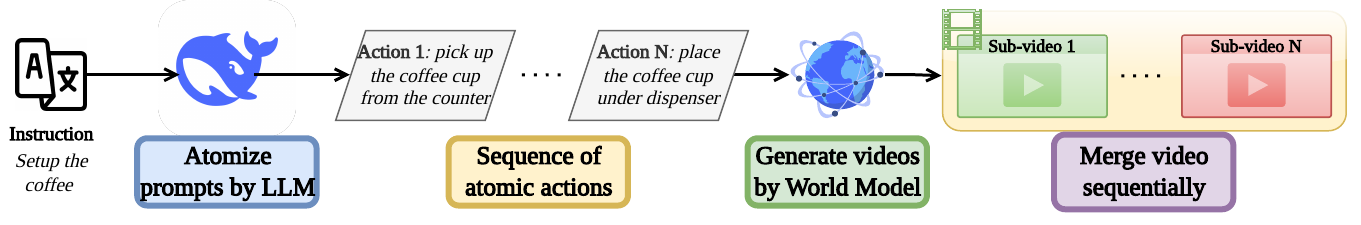}
    \caption{\textbf{Post-training rollout synthesis.} 
    Given an initial frame and a high-level instruction, an LLM decomposes the instruction into an ordered list of atomic skill prompts. 
    The world model generates one video segment per skill, using the final frame of each segment as the visual condition for the next skill. 
    The resulting video segments are concatenated to form a long-horizon synthetic rollout. 
    We retain the pre-decoding video latents from the world model as post-training supervision for \textsc{World2Act}. 
    The atomic-skill data used to train the world model are described in~\autoref{appendix-sec:data-wm}.}
    \label{fig:post_training_synthesis}
\end{figure}

We use a Cosmos-Predict2 world model~\citep{agarwal2025cosmos}, finetuned on atomic-skill datasets as described in~\autoref{appendix-sec:data-wm}, to provide target latent rollouts for Stage~2 post-training of \textsc{World2Act}. For each of the RoboCasa, LIBERO, and Bridge-SIMPLER benchmarks, we sample 1,000 scene initializations, each consisting of an initial observation and a language instruction. This yields roughly 3,000 post-training rollouts in total. We ensure that the post-training scene initializations are disjoint from the evaluation scenes to prevent test leakage. The synthesis procedure is skill-compositional, as illustrated in~\autoref{fig:post_training_synthesis}.

During post-training, the WM imagination branch and VLA control branch run in parallel from the same initial scene and instruction. In the WM branch, we first prompt DeepSeek~\citep{Guo2025DeepSeekR1} to decompose the high-level instruction into an ordered sequence of atomic skill descriptions defined in a manually specified schema. Each atomic skill description specifies a short-horizon manipulation primitive, such as picking up an object, moving it to a target location, or placing it under a dispenser. Cosmos-Predict2 then generates one short video segment for each atomic skill prompt, autoregressively conditioning each segment on the final frame of the previous one. The resulting segments are concatenated into a long-horizon imagined rollout, and we retain the corresponding pre-decoding video latents as the WM target trajectory. In the VLA branch, the current augmented policy is rolled out from the same initial scene. \textsc{World2Act} post-trains the residual policy by aligning the policy action-latent trajectory with the parallel WM target latent trajectory, without using decoded pixels, simulator rewards, or expert action labels.

For analyses that require ground-truth action targets, such as Fig.~\ref{fig:opendrawer_tsne} and Table~\ref{tab:action_mse}, we separately use the \textit{Human} split of RoboCasa, which provides expert action annotations. We sample 50 scenes from this split and report the mean-squared error averaged across these scenes. This evaluation set is used only for action-space analysis and is distinct from the synthetic rollouts used for post-training.

\subsection{Cross-Task Generalization}
\label{appendix-subsec:cross-task-generalization}

\autoref{tab:cross_task_split} presents the train-test split for the cross-task generalization experiment (Fig.~\ref{fig:generalization_seen_tasks} of the main paper). We designate \textit{Pick and Place} (PnP) as the primary held-out task suite for evaluating zero-shot transfer under a significant distributional shift. While PnP is never observed as an explicit task during post-training, the robot frequently encounters directly relevant interaction skills, for example, reaching, grasping, lifting, and placement-like motions, which are implicitly learned through related tasks like \textit{Open Door} or \textit{Turn Off Microwave}. Our experiment is designed to reuse these transferable skills to synthesize behavior in novel task compositions, making PnP a learnable but non-trivial generalization target. At $12$ seen training tasks, \textsc{World2Act} improves cross-task success rate over the corresponding vanilla policies by $2.4\%$ for GR00T-N1.6-base and $1.1\%$ for Cosmos Policy.

\begin{table}[!ht]
    \caption{\textbf{Cross-task generalization experiment splits.}}
    \label{tab:cross_task_split}
    \centering
    \begin{tabular}{ll}
        \toprule
        \textbf{Train Tasks} & \textbf{Test Tasks} \\
        \midrule
        CoffeeSetupMug & PnPCounterToCab \\
        CoffeeServeMug & PnPCabToCounter \\
        CoffeePressButton & PnPCounterToSink \\
        OpenSingleDoor & PnPSinkToCounter \\
        OpenDoubleDoor & PnPCounterToMicrowave \\
        CloseSingleDoor & PnPMicrowaveToCounter \\
        CloseDoubleDoor & PnPCounterToStove \\
        OpenDrawer & PnPStoveToCounter \\
        CloseDrawer & TurnOnSinkFaucet \\
        TurnOnMicrowave & TurnOffSinkFaucet \\
        TurnOffMicrowave & TurnSinkSpout \\
        TurnOffStove & TurnOnStove \\
        \bottomrule
    \end{tabular}
\end{table}

\subsection{Real-World Experiments}
\label{appendix-subsec:real-world-experiments}

To evaluate our method's sim-to-real transfer capabilities, our physical evaluation suite is designed to replicate simulation tasks from the RoboCasa and LIBERO benchmarks closely. We define success for the real-world tasks as follows: (i) \textit{Pick and place cup}: The robot must pick the cup and place it stably on the plate in a single, continuous attempt. (ii) \textit{Pick up bowl}: The robot must secure the bowl and return to its home configuration (see~\autoref{fig:robot_setup} in the main paper) while maintaining a stable lift. (iii) \textit{Close microwave}: A trial is successful when the microwave door is pushed closed, and its magnet fully engages. We provide additional demonstrations in our video to illustrate world-model rollouts in the real-world Franka setup, highlighting the execution of our approach across three distinct physical tasks, demonstrating robust transfer from the WM to real-world control.

An important detail for the \textit{Pick and place cup} task is our use of atomic-skill segmentation, which aligns with our post-training data construction to ensure consistency during transfer from WM to VLA. The task is segmented into two atomic skills: ``pick cup'' and ``place cup.'' To count as a single success, the robot must complete both skills in a single shot. When training the WM and our \textsc{World2Act}, we do not train the full pick-and-place behavior as a one-shot sequence; instead, we collect demonstrations for the \textit{Pick} and \textit{Place} skills separately and train them as two atomic skills.

\subsection{Additional Latent Visualizations}
\label{app:extra_latent_visualization}

\begin{figure}[t]
    \centering

    \includegraphics[width=\linewidth]{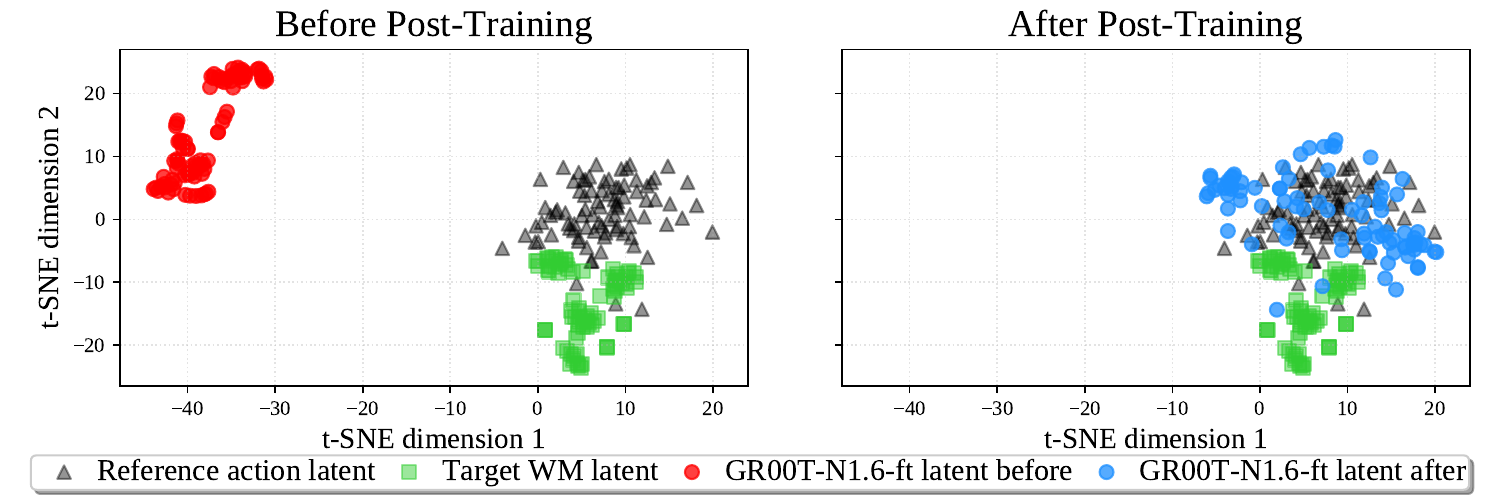}
    \vspace{-0.5em}
    \centerline{\small (a) {PnPCounterToCab}}

    \vspace{0.75em}

    \includegraphics[width=\linewidth]{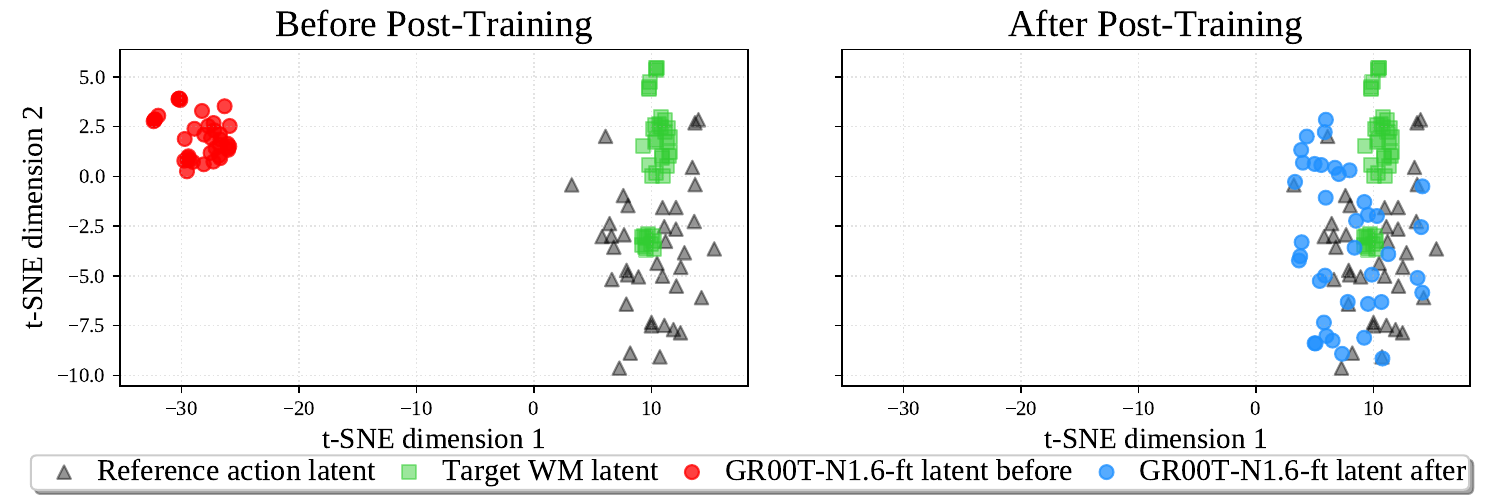}
    \vspace{-0.5em}
    \centerline{\small (b) {TurnOnStove}}

    \caption{
    \textbf{Additional t-SNE visualizations.} We provide extra qualitative results  on {PnPCounterToCab} and {TurnOnStove}. Consistent with the \textit{OpenDrawer} results in~\autoref{fig:opendrawer_tsne}, \textsc{World2Act} yields predicted-action  latent distributions that better overlap with the reference-action and target-WM latent regions, indicating improved alignment while remaining compatible with the expert action manifold.
    }
    \label{fig:extra_tsne}
\end{figure}

We additionally provide t-SNE visualizations for two more tasks, \textit{PnPCounterToCab} and \textit{TurnOnStove}, complementing the \textit{OpenDrawer} analysis in~\autoref{fig:opendrawer_tsne}. We follow the same visualization protocol as in the main paper: reference expert actions correspond to ground-truth actions from held-out demonstrations, target-WM latents correspond to the target world-model representations, and predicted-action latents are produced by the learned policy. For each task, we sample $50$ held-out scenes from the \textit{Human} split of RoboCasa, which provides expert action annotations. As shown in~\autoref{fig:extra_tsne}, \textsc{World2Act} produces predicted-action latents that are more consistently co-located with the reference-action and target-WM latent regions than the baseline. These additional visualizations suggest that the latent alignment observed on \textit{OpenDrawer} is not task-specific, and that \textsc{World2Act} preserves compatibility with the expert action manifold across manipulation tasks with different object interactions and scene dynamics.

\subsection{Inference Speed Analysis}
\label{appendix-subsec:inference-speed}

\begin{table}[!ht]
  \caption{\textbf{Inference speed comparison.} We report the inference speed on RoboCasa using an NVIDIA RTX 4090. We measure speed by the time required to predict an action from an observation.}
  \label{tab:inference_speed}
  \centering
  \begin{tabular}{@{}lcc@{}}
    \toprule
    Model & Inference Speed (Hz) ($\uparrow$) & Per Action (ms) ($\downarrow$) \\
    \midrule
    \rowcolor{grootgreen} GR00T-N1.6-ft~\cite{nvidia_gear_gr00t_n1_6_2025} & 274.1 & 3.6 \\
    \rowcolor{grootgreen} GR00T-N1.6-ft + \textsc{World2Act} (Ours) & 251.9 & 4.0 \\
    \midrule
    \rowcolor{cosmosblue} Cosmos Policy~\cite{kim2026cosmos} & 20.8 & 48.1 \\
    \rowcolor{cosmosblue} Cosmos Policy + \textsc{World2Act} (Ours) & 20.5 & 48.8 \\
    \bottomrule
  \end{tabular}
\end{table}

\autoref{tab:inference_speed} compares the inference latency of the baseline VLAs against those integrated with our \textsc{World2Act}. The results show that our method adds minimal computational overhead. When applied to GR00T-N1.6-ft, \textsc{World2Act} maintains a rate above 250 Hz. This speed facilitates physical applications, as demonstrated by our robot experiments in~\autoref{subsec: real-robot} of the main paper. Cosmos Policy operates at a lower rate of about 20.8 Hz per step because it uses a joint representation for video, language, and action. Even so, including \textsc{World2Act} drops the speed of Cosmos Policy by just 1.5\%. This high-dimensional joint latent space likely accounts for the incremental performance gains observed with Cosmos Policy in the~\autoref{appendix-sec:libero-results}, as the model's high-dimensional embeddings already capture significant cross-modal correlations. 

\begin{table}[!ht]
\caption{\textbf{Architectural details of the policy components.} In our implementation, $A=12$ for RoboCasa and $A=7$ for LIBERO, Bridge-SIMPLER and real Franka Robot, $M=4$, $D=32$, and $D_{\text{proprio}}=53$. Input images are $256 \times 256$. The Attention Module processes $N_{\text{src}}=3$ tokens: the visual feature, the latent base action from $\mathcal{B}_{\text{a}}$, and the proprioception state. The video latent resolution is $16\times60\times104$, following the Cosmos-Predict-2~\cite{agarwal2025cosmos} implementation.}
\label{tab:architecture_details}
\centering
\resizebox{\linewidth}{!}{
\begin{tabular}{lll}
\toprule
\textbf{Module} & \textbf{Tensor Mapping} & \textbf{Architecture} \\
\midrule
\textbf{Action Adapter} $\mathcal{B}_{\text{a}}$ & $\mathbb{R}^{A \times M} \rightarrow \mathbb{R}^{D}$ & Flatten, FC($A \times M$, 128), GELU \\
(Action Encoder) & & FC(128, 64), GELU \\
 & & FC(64, $D$) \\
\midrule
\textbf{Video Adapter} $\mathcal{B}_{\text{v}}$ & $\mathbb{R}^{C \times H \times W} \rightarrow \mathbb{R}^{D}$ & Conv2D($C$, 64, $k=3, s=2, p=1$), GroupNorm, GELU \\
(Video Encoder) & & Conv2D(64, 128, $k=3, s=2, p=1$), GroupNorm, GELU \\
 & & AdaptiveAvgPool2D((1, 1)), Flatten, FC(128, $D$) \\
\midrule
\textbf{Action Decoder} $\mathcal{D}_{\text{a}}$ & $\mathbb{R}^{D} \rightarrow \mathbb{R}^{A \times M}$ & FC($D$, 64), GELU \\
& & FC(64, 128), GELU \\
 & & FC(128, $A \times M$) \\
\midrule
\textbf{Residual Network} $f^{\theta}$ & & \\
\quad \textit{State Encoder} & $\mathbb{R}^{D_{\text{proprio}}} \rightarrow \mathbb{R}^{D}$ & FC($D_{\text{proprio}}$, 128), ReLU, FC(128, $D$), LayerNorm \\
\quad \textit{Visual Encoder} & $\mathbb{R}^{3 \times H_{img} \times W_{img}} \rightarrow \mathbb{R}^{D}$ & Conv2D(3, 16, $k=8, s=4, p=0$), ReLU \\
 & & Conv2D(16, 32, $k=4, s=2, p=0$), ReLU \\
 & & Conv2D(32, 32, $k=3, s=2, p=0$) \\
 & & Flatten, FC(6272, $D$), LayerNorm \\
\quad \textit{Transformer Module} & $\mathbb{R}^{N_{\text{src}} \times D} \rightarrow \mathbb{R}^{N_{\text{src}} \times D}$ & Self-Attention (2 layers, 4 heads, $d_{\text{model}}=D$) \\
\bottomrule
\end{tabular}
}
\end{table}

\section{Network Specifications}
\label{appendix-sec:network-specifications}

Following the notation established in the main paper, we provide the detailed architectural specifications for the action adapter $\mathcal{B}_{\text{a}}$, video adapter $\mathcal{B}_{\text{v}}$, action decoder $\mathcal{D}_{\text{a}}$, and the residual network $f^{\theta}$ in~\autoref{tab:architecture_details}. As defined previously, $M$ denotes the action chunk size, $D$ the latent hidden dimension, and $C \times H \times W$ the resolution of the video latents per chunk. Additionally, we define $D_{\text{proprio}}$ as the robot proprioception state dimension, $A$ as the action dimension, and $3 \times H_{\text{img}} \times W_{\text{img}}$ as the input image resolution. For convolutional layers, parameters are denoted as (input\_channels, output\_channels, $k$=kernel\_size, $s$=stride, $p$=padding). For fully connected layers, we use FC(input\_features, output\_features).

\paragraph{Residual Policy Architecture.}
We provide the architectural details of the residual policy used in~\hyperref[sub-sec:Stage-2]{Stage~2}. At each chunk boundary, the frozen VLA \(\pi_{\mathrm{base}}\) predicts a base action chunk \(\bar{a}_{\mathrm{base},t}\) from the current state \(s_t\) and instruction \(\ell\):
\begin{equation}
    \bar{a}_{\mathrm{base},t}
    =
    \pi_{\mathrm{base}}(s_t,\ell).
\end{equation}
Since residual corrections are learned in the Stage-1 action-latent space, we embed this base action chunk using the frozen action adapter \(\mathcal{B}_{\mathrm a}\) to obtain the base action-latent token:
\begin{equation}
    \mathbf{x}^{(0)}
    =
    z^{\mathrm a}_{\mathrm{base},t}
    =
    \mathcal{B}_{\mathrm a}(\bar{a}_{\mathrm{base},t}).
\end{equation}

The residual policy \(f^\theta\) is conditioned on the current state \(s_t\) and the base action latent \(z^{\mathrm a}_{\mathrm{base},t}\). We encode visual observations in \(s_t\) using a CNN and robot proprioceptive observations using an MLP, yielding a set of observation tokens \(\{\mathbf{x}^{(1)}, \dots, \mathbf{x}^{(n)}\}\). We concatenate the action-latent token with the observation tokens, \(\{\mathbf{x}^{(0)}, \mathbf{x}^{(1)}, \dots, \mathbf{x}^{(n)}\}\), and process the resulting token sequence with a self-attention Transformer, producing contextualized features \(\{\mathbf{h}^{(0)}, \mathbf{h}^{(1)}, \dots, \mathbf{h}^{(n)}\}\). An MLP head applied to the contextualized action token \(\mathbf{h}^{(0)}\) predicts the latent residual:
\begin{equation}
    \Delta z^{\mathrm a}_t
    =
    \mathrm{MLP}(\mathbf{h}^{(0)}).
\end{equation}

The refined action latent is computed as
\begin{equation}
    z^{\mathrm a}_{\mathrm{final},t}
    =
    z^{\mathrm a}_{\mathrm{base},t}
    +
    \Delta z^{\mathrm a}_t,
\end{equation}
and decoded by the frozen action decoder:
\begin{equation}
    \bar{a}_{\mathrm{final},t}
    =
    \mathcal{D}_{\mathrm a}(z^{\mathrm a}_{\mathrm{final},t}).
\end{equation}

During Stage~2, only the parameters of \(f^\theta\) are updated; the base VLA, action adapter \(\mathcal{B}_{\mathrm a}\), video adapter \(\mathcal{B}_{\mathrm v}\), and action decoder \(\mathcal{D}_{\mathrm a}\) remain frozen.

\section{Data for World Model}
\label{appendix-sec:data-wm}
\label{appendix-sec:data-details}

Current video world models struggle with long-video generation because they are commonly trained on fixed-length clips, while robotic executions vary widely in duration~\cite{huang2026live}. This mismatch can make training unstable for long-horizon manipulation trajectories, where different tasks may contain substantially different temporal structures. To reduce this variance, we use an automatic atomic-skill segmentation pipeline that decomposes demonstrations into shorter action segments paired with low-level language prompts. This increases the uniformity of world-model training videos and provides temporally focused clips for learning robot-object interaction dynamics. This section details the data decomposition pipeline, validates the resulting datasets, and describes the additional formatting used for world-model training.

\begin{figure}[!ht]
    \centering
    \includegraphics[width=\linewidth, trim={0 0 0 0.1cm}, clip]{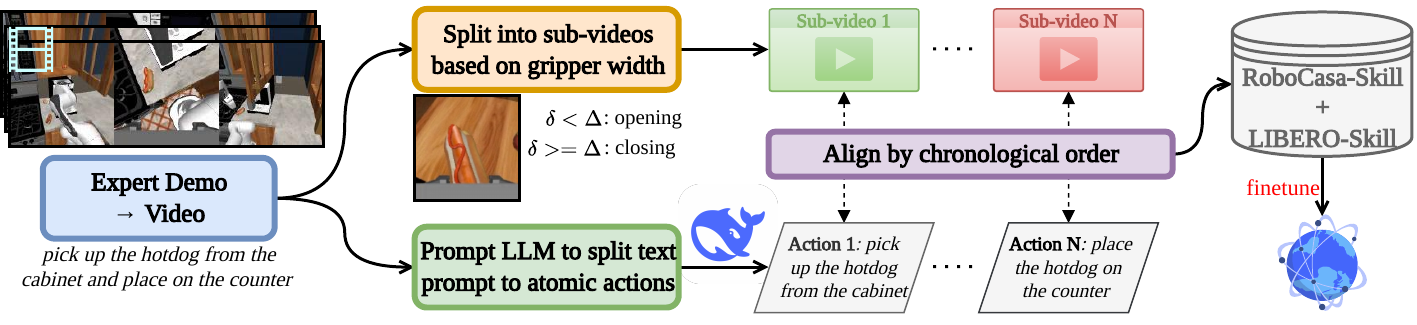}
    \caption{\textbf{Atomic-skill data construction for world-model training.} We segment each demonstration using gripper-state changes, decompose the global instruction into atomic skill prompts with an LLM, and synchronize sub-videos with prompts through schema-based chronological alignment.}
    \label{fig:data_pipeline}
\end{figure}

\subsection{Data Processing and Synchronization}
\label{appendix-subsec:data-pipeline}

\autoref{fig:data_pipeline} summarizes our atomic-skill data construction pipeline. Given a long-horizon demonstration, we first identify candidate interaction boundaries from the gripper aperture. Let \(w_t\) denote the gripper width at time \(t\), and let \(w_0\) denote the calibrated fully open width at the start of an episode. We define the closure signal as \(\delta_t = w_0 - w_t\), where \(\delta_t \approx 0\) indicates that the gripper remains fully open, while larger values indicate increasing closure. Each frame is classified into an event \(E_t \in \{\text{\textit{contact}}, \text{\textit{non-contact}}\}\), assigning \(E_t = \text{\textit{contact}}\) if \(\delta_t \ge \Delta\), and \(E_t = \text{\textit{non-contact}}\) otherwise. Candidate sub-video segments are then constructed around complete action cycles, starting from \textit{non-contact} frames and ending at the completion of a corresponding \textit{contact} event.

To assign language labels to these visual segments, we use DeepSeek~\cite{Guo2025DeepSeekR1} to decompose each global, high-level instruction into an ordered sequence of atomic skill prompts. We first manually define task schemas for each simulator. These schemas provide a standardized skill vocabulary for each task family, ensuring consistent video-language synchronization across environments. Given a global instruction, a task schema, and a set of \(\ell\) detected contact indices, the LLM aligns valid indices to schema steps in chronological order. Ideally, \(\ell\) matches the number of steps in the corresponding schema; for example, a ``pick and place'' task requires two steps: ``pick'', then ``place''.

Because raw contact indices can contain noise, such as duplicate triggers or temporally clustered detections, we explicitly instruct the LLM to identify and filter out false positives. The model aligns the remaining valid indices chronologically with the schema steps. To prevent hallucinated frame numbers, the LLM is instructed to use only the provided indices. If an execution contains fewer valid indices than required by the schema, the LLM flags the missing steps, and we discard the incomplete sequence. The remaining valid indices define the final sub-video boundaries, and their aligned schema steps provide the corresponding low-level prompts. This produces short-horizon sub-videos paired with atomic language commands for world-model training.

\paragraph{LLM Prompt Template.}
The system prompt used for schema-based skill alignment is shown below. We ask the model to output strict JSON, provide a brief explanation for its filtering and alignment decisions, and return the final mapping from valid contact indices to atomic skill labels.

\begin{tcolorbox}[colback=gray!5,colframe=gray!50,arc=4pt,boxrule=1pt,left=5pt,right=5pt,top=5pt,bottom=5pt,breakable]
\small
\begin{verbatim}
You are a robotic video dataset processing assistant. You output strict 
JSON.

OBJECTIVE:
You will receive a video task with detected timestamp indices. Your 
goal is to align these indices to the correct schema steps, filtering 
out sensor noise.

LOGIC FLOW:
1. ANALYZE: Compare the User Caption against the Detected Indices.
2. FILTER: Identify indices that are likely NOISE, HESITATION, or 
DUPLICATES.
3. ALIGN: Map the Schema Steps to the valid indices chronologically.
4. HANDLE MISSING: 
   - If there are FEWER indices than steps, flag it. 
   - If there are MORE indices than steps, mark extras as NOISE.
   - DO NOT invent new frame numbers.

OUTPUT FORMAT:
Return a JSON object with a "reasoning" field and an "alignment" field.

EXAMPLE:
User Input: 
Task: "CoffeeSetupMug", Indices: [116, 230, 235], Schema: ["pick", 
"place"], 
Caption: "Pick the mug and place it under the coffee dispenser."
Output:
{
  "reasoning": "The caption confirms both pick and place actions. Index 
  116 aligns with `pick'. Indices 230 and 235 are temporally close; 
  235 is filtered as sensor noise during the `place' execution.",
  "alignment": {
    "116": "pick(mug)",
    "230": "place(mug)"
  }
}
\end{verbatim}
\end{tcolorbox}

For each demonstration, we input the global instruction, task schema, and detected contact indices into the prompt. We then extract the resulting alignment from the model output and use the aligned schema steps as atomic skill labels for the processed world-model training datasets.

\subsection{Validating Atomic-Skill Decomposition}
\label{appendix-subsec:data-evaluation}

Accurate pairing between video segments and prompts is essential for training a reliable world model. We therefore evaluate both cross-modal synchronization quality and the resulting distribution of video lengths. The goal is to verify that the decomposition process produces well-aligned video-language pairs while mitigating the long-tail duration issues present in the original demonstrations.

\begin{figure}[!ht]
    \centering

    \begin{subfigure}[t]{0.49\linewidth}
    \centering
    \includegraphics[width=\linewidth, trim={0 0 0 0.7cm}, clip]{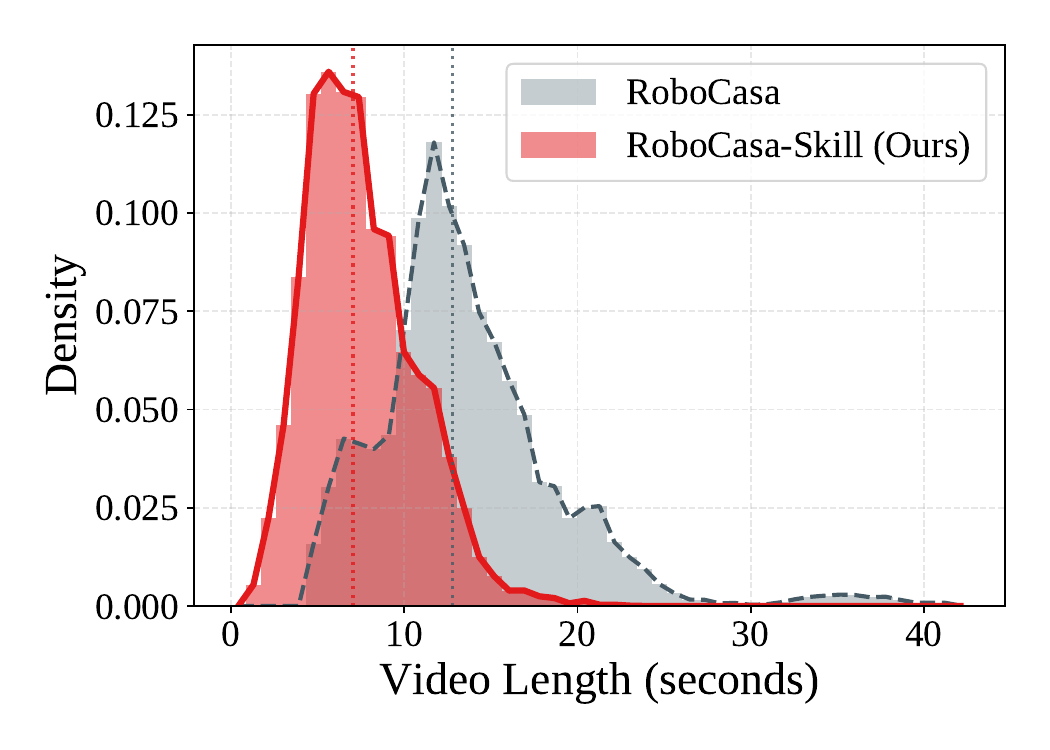}
    \caption{\textbf{RoboCasa.}}
    \label{fig:tasklen-robocasa}
    \end{subfigure}\hfill
    \begin{subfigure}[t]{0.49\linewidth}
    \centering
    \includegraphics[width=\linewidth, trim={0 0 0 0.7cm}, clip]{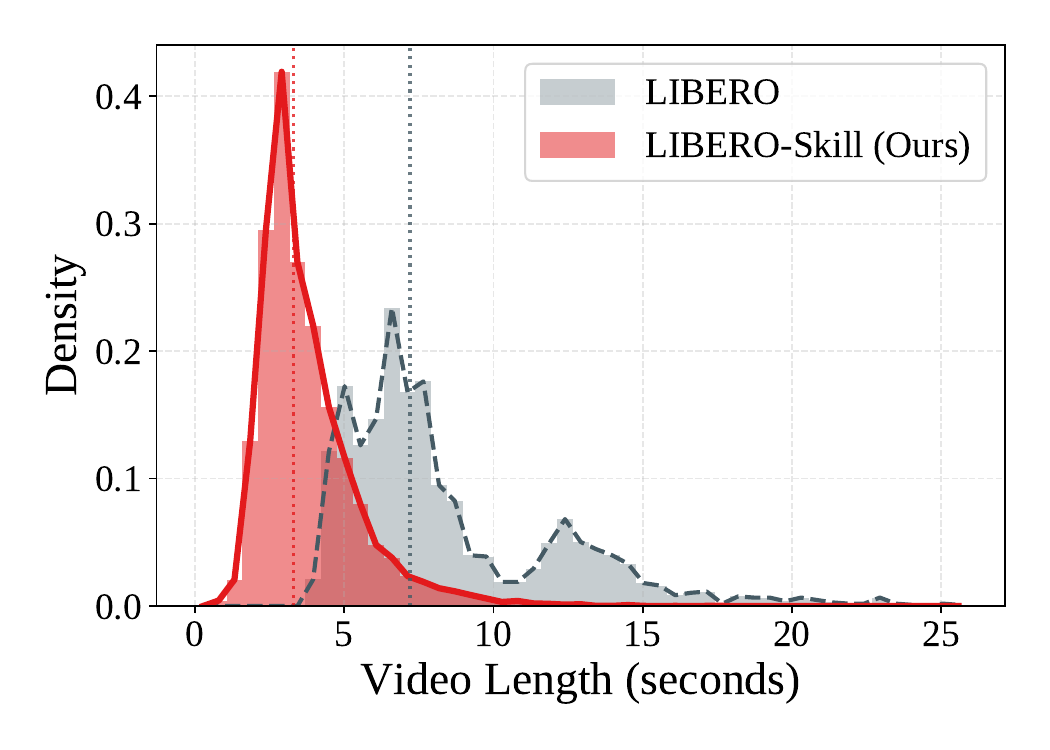}
    \caption{\textbf{LIBERO.}}
    \label{fig:tasklen-libero}
    \end{subfigure}

    \caption{\textbf{Video-length distribution.} Across RoboCasa and LIBERO, decomposed skills exhibit shorter mean length and a more concentrated distribution than full sequences, improving stability for world-model training on variable-duration robotic demonstrations.}
    \label{fig:tasklen-sidebyside}
\end{figure}

Our pipeline achieves synchronization rates, defined as the successful one-to-one matching between segmented sub-videos and generated atomic prompts, of \(96.2\%\) on the processed RoboCasa data and \(86.9\%\) on the processed LIBERO data. These results demonstrate high alignment rates between visual segments and language labels. Infrequent failures are largely due to rare gripper noise or non-prehensile interactions~\cite{shirai2026learning}, such as using the arm rather than the gripper to close a door.

\autoref{fig:tasklen-sidebyside} shows that skill decomposition reshapes the original high-variance, long-tailed video-length distributions (\textit{gray}) into more concentrated, approximately unimodal distributions (\textit{red}). Specifically, the density around the median increases by \(17\%\) for the processed RoboCasa data and \(72\%\) for the processed LIBERO data, indicating that a much larger fraction of training clips fall within a uniform temporal duration. This mass concentration yields a more consistent prediction horizon, reducing distributional variance and stabilizing world-model training under variable task durations.

\subsection{Dataset Statistics}
\label{appendix-subsec:dataset-statistics}

Using the optimized closure threshold \(\Delta = 0.005\textrm{m}\), our processed RoboCasa dataset contains 114,192 video sequences, compared to 67,593 sequences in the original RoboCasa dataset. For LIBERO, our processed dataset contains 11,782 video sequences, compared to 2,007 sequences in the original {LIBERO} dataset. This growth is a direct consequence of the skill-based segmentation pipeline, which extracts multiple atomic skill segments from each long-horizon trajectory. The resulting datasets provide substantially more short-horizon video-language pairs for world-model training while preserving chronological consistency with the original demonstrations. For {Bridge-SIMPLER} simulator, the expert demonstration videos have relatively uniform lengths, so we directly use the simulator's original expert demonstrations without additional segmentation. 

\subsection{Multi-view Stitching}
\label{appendix-subsec:multiview-stitching}

\begin{figure}[!ht]
    \centering
    \begin{subfigure}[b]{0.48\textwidth}
        \centering
        \includegraphics[width=\linewidth]{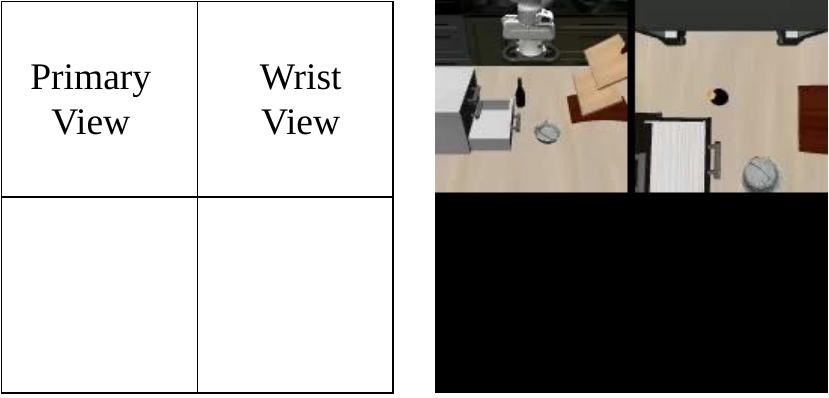}
        \caption{LIBERO layout and rendered sample.}
        \label{fig:stitching_libero}
    \end{subfigure}
    \hfill
    \begin{subfigure}[b]{0.48\textwidth}
        \centering
        \includegraphics[width=\linewidth]{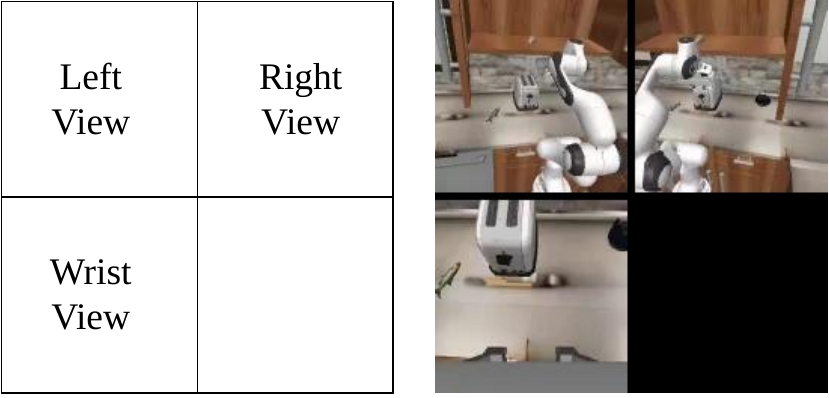}
        \caption{RoboCasa layout and rendered sample.}
        \label{fig:stitching_robocasa}
    \end{subfigure}
    \caption{\textbf{Multi-view stitching strategies.} To preserve geometric context, observations from multiple cameras are spatially arranged into a unified \(2 \times 2\) grid. For each simulation environment, we illustrate the camera mapping template on the left and an example stitched observation on the right. \textbf{(a)} In LIBERO, the primary and wrist cameras occupy the top row, with the bottom row black-padded. \textbf{(b)} In RoboCasa, the left, right, and wrist views are mapped to the top-left, top-right, and bottom-left quadrants, respectively.}
    \label{fig:multiview_stitching}
\end{figure}

Following Cosmos-Predict2~\cite{agarwal2025cosmos}, we stitch multi-view observations into a single composite layout for each video frame to enforce geometric consistency and enhance the model's spatial understanding. Specifically, the RGB observations are arranged into a \(2 \times 2\) grid layout. For RoboCasa, the grid consists of the left, right, and wrist camera views in the top-left, top-right, and bottom-left panes, respectively, with the bottom-right pane left black. For LIBERO, the primary and wrist camera views occupy the top-left and top-right panes, while the entire bottom row remains black. \autoref{fig:multiview_stitching} illustrates these configurations.

\section{World Model Training and Evaluation}
\label{appendix-sec:wm-evaluation}
\label{appendix-sec:wm-selection}

\subsection{Backbones}
We finetune four recent, high-performing video diffusion backbones: {LTX-Video}~\cite{hacohen2024ltx}, {HunyuanVideo-1.5}~\cite{hunyuanvideo2025}, {Wan2.2}~\cite{wan2025}, and {Cosmos-Predict2}~\cite{agarwal2025cosmos}. Each model is trained for 10K steps on our processed datasets. To ensure a fair comparison, we follow the model sizes and training configurations recommended by the original authors. Specifically, most models adopt LoRA as the standard finetuning strategy, whereas {Cosmos-Predict2} uses full finetuning for its 2B variant; we therefore follow this protocol. For the LoRA-based models, we use the 13B version of {LTX-Video} with rank 128, the 5B version of {Wan2.2} with rank 32, and the 480P-I2V configuration of {HunyuanVideo-1.5} with rank 8.

We additionally train a {Cosmos-Predict2} model on the original RoboCasa dataset using the standard \textit{MG} split~\cite{nasiriany2024robocasa}. We refer to this model as Base-WM, and denote the variant trained on our processed atomic datasets as Skill-WM. All models are trained in a distributed setup using eight AMD Instinct MI210 GPUs, each with 64~GB of VRAM.

\subsection{Evaluation Protocol}
\label{appendix-subsec:wm-evaluation-protocol}

We evaluate WM backbone generative quality following the protocol in~\cite{jang2025dreamgen} using 100 representative Pick and Place trajectories in RoboCasa. We evaluate all baselines on the \textit{Human} split, a held-out set disjoint from the \textit{MG} split~\cite{nasiriany2024robocasa}, which is used for finetuning. We assess performance using two key metrics: \textbf{Instruction Following (IF)}, measured via Qwen-VL-2.5~\cite{bai2025qwen2} to check alignment between text prompts and visual results, and \textbf{Physical Alignment (PA)}, measured via VideoCon-Physics~\cite{bansal2025videophy} to evaluate the consistency of physical interactions in generated videos.

\subsection{Quantitative Results}
\label{appendix-subsec:wm-quantitative}

\autoref{tab:backbone_results} shows Cosmos-Predict2 achieves the best IF and PA performance thanks to its pretraining on extensive robotic datasets that provide more robust spatiotemporal priors for manipulation tasks compared to other generic backbones~\cite{agarwal2025cosmos}. Consequently, we choose it as the main backbone throughout the paper. Furthermore, training Cosmos-Predict2 on our atomic-skill-segmented data significantly improves both IF and PA metrics compared to the base backbone. We attribute this to the distribution of the training data; standard trajectories in robotics datasets often vary in duration, biasing toward simpler, shorter sequences while ignoring longer-horizon videos. By atomizing data into discrete skills, the \textbf{length distribution during training becomes more uniform} (as visualized in~\autoref{fig:tasklen-sidebyside}). This allows the model to learn robust transitions and complex contact dynamics without being overwhelmed by the temporal noise of long-horizon trajectories.

\begin{table}[!ht]
  \caption{\textbf{Quantitative comparison of video diffusion backbones.} The Cosmos-Predict2 model trained on our atomic-skill data significantly outperforms baselines in both instruction following and physical alignment. Our results are consistent with Jang \textit{et al.}~\cite{jang2025dreamgen}.}
  \label{tab:backbone_results}
  \centering
  \begin{tabular}{@{}lcc@{}}
    \toprule
    Video Diffusion Backbone & IF ($\uparrow$) & PA ($\uparrow$) \\
    \midrule
    LTX-Video~\cite{hacohen2024ltx} & 7.7 & 40.1 \\
    HunyuanVideo-1.5~\cite{hunyuanvideo2025} & 8.3 & 44.8 \\
    Wan2.2~\cite{wan2025} & 20.5 & 56.9 \\
    Cosmos-Predict2~\cite{agarwal2025cosmos} (Base data) & 29.6 & 60.5 \\
    \rowcolor{cosmosblue} Cosmos-Predict2 (Ours, atomic-skill data) & \textbf{35.3} & \textbf{65.4} \\
    \bottomrule
  \end{tabular}
\end{table}

\subsection{Qualitative Analysis: Where Atomic-Skill Segmentation Matters}
\label{appendix-subsec:wm-qualitative}
\begin{figure}[!ht]
    \centering
    \includegraphics[width=0.9\linewidth, trim={0 5pt 0pt 4pt}, clip]{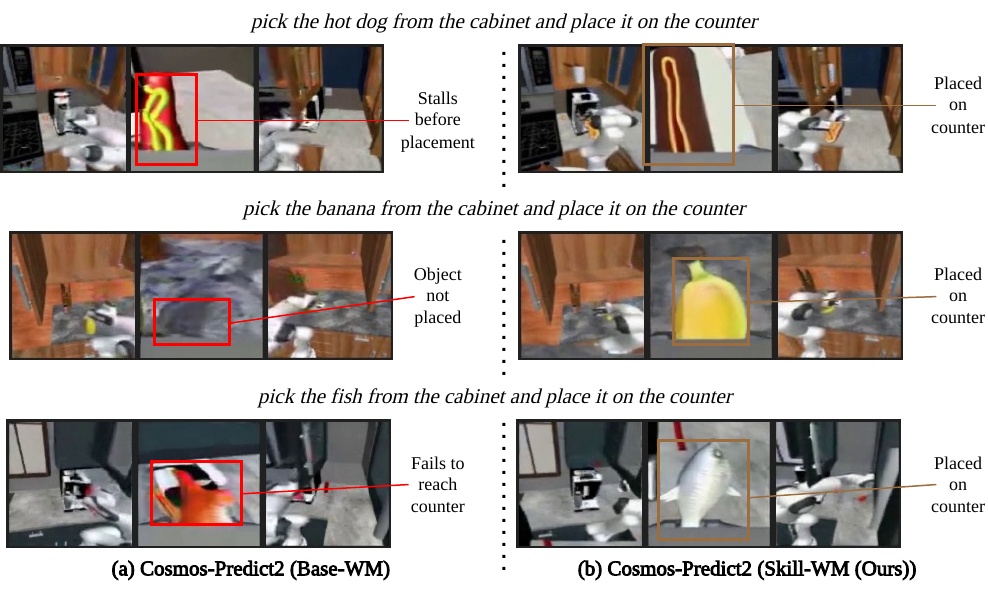}
    \caption{\textbf{Qualitative comparison of task completion: Skill-WM vs. Base-WM.} We display the final frames of execution videos generated from the same prompts and initial states. While the Base-WM (left) fails to complete the trajectory and stalls mid-task, the WM trained on atomic-skill data (right) successfully places the object on the counter. Each example depicts three synchronized camera perspectives: left, wrist-mounted, and right.}
    \label{fig:quallitative-wm}
\end{figure}

Our qualitative study highlights the critical role of temporal decomposition in multi-step tasks. As illustrated by the final frames in~\autoref{fig:quallitative-wm}, the Base-WM frequently encounters ``mid-task'' failure modes, in which the generation stalls or fails to ground the target object. This instability is a direct result of accumulated error when the base model attempts to generate excessively long video sequences without explicit grounding of atomic skills. In contrast, the WM trained on atomic-skill data maintains structural and temporal coherence by learning from shorter, manageable segments. This approach alleviates the drift issues inherent to long-horizon generation, enabling the model to complete the \textit{Pick and Place} tasks.

\subsection{Skill-Compositional World Models Improve Downstream Success}

\begin{table}[!ht]
    \centering
    \small
    \caption{
        \textbf{Impact of WM on \textsc{World2Act}.}
        Our Skill-WM improves downstream success rate across both policies.
    }
    \label{tab:wm-comparison}\renewcommand{\arraystretch}{1.15}
    \begin{tabular}{l l r}
        \toprule
        \textbf{Method} & \textbf{World Model} & \textbf{SR} \\
        \midrule

        \rowcolor{cosmosblue}
        Cosmos Policy~\cite{kim2026cosmos} + \textsc{World2Act}
        & Base-WM & 0.661 \\
        \rowcolor{cosmosblue}
        & Skill-WM (Ours) & \textbf{0.663} \\

        \midrule

        \rowcolor{grootgreen}
        GR00T-N1.6-ft~\cite{nvidia_gear_gr00t_n1_6_2025} + \textsc{World2Act}
        & Base-WM & 0.715 \\
        \rowcolor{grootgreen}
        & Skill-WM (Ours) & \textbf{0.726} \\

        \bottomrule
    \end{tabular}
\end{table}

\autoref{tab:wm-comparison} shows
Skill-WM consistently improves downstream success because it is trained on skill-compositional data, producing more temporally consistent rollouts. This better matches \textsc{World2Act}'s chunk-wise alignment in a shared video--action space, leading to stronger policy execution.

\section{Failure Cases Analysis}
\label{appendix-sec:failure-case}

\subsection{Imperfect World Model Rollouts}
\label{appendix-subsec:wm-rollout-failures}

\begin{figure}[!ht]
    \centering
    \setlength{\fboxsep}{0pt}
    \setlength{\fboxrule}{0.5pt}

    \begin{subfigure}[b]{0.45\linewidth}
        \centering
        \fbox{\includegraphics[width=\textwidth]{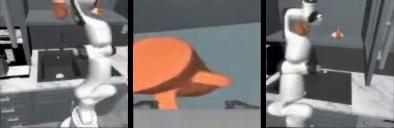}}
        \caption{RoboCasa}
    \end{subfigure}
    \hfill
    \begin{subfigure}[b]{0.30\linewidth}
        \centering
        \fbox{\includegraphics[width=\textwidth]{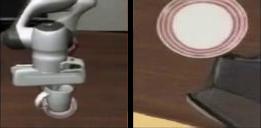}}
        \caption{LIBERO}
    \end{subfigure}
    \hfill
    \begin{subfigure}[b]{0.15\linewidth}
        \centering
        \fbox{\includegraphics[width=\textwidth]{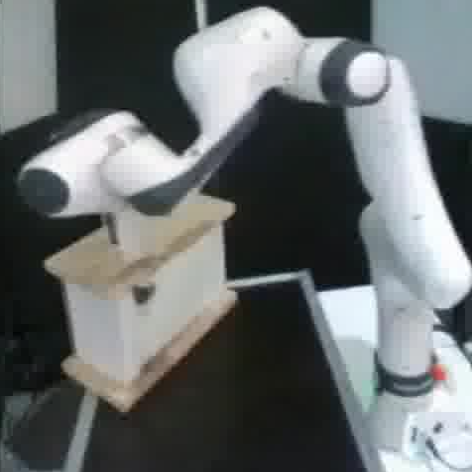}}
        \caption{Real}
    \end{subfigure}

    \caption{\textbf{Visual artifacts examples.} We illustrate representative failure cases across different environments. The RoboCasa, LIBERO, and real-world setups utilize three, two, and one camera views, respectively.}
    \label{fig:hallucination_analysis}
\end{figure}

While our skill-compositional WM demonstrates temporally consistent video generation, it occasionally yields imperfect rollouts, which is a known limitation of current video diffusion models~\cite{kim2026cosmos}. We illustrate representative failure cases in~\autoref{fig:hallucination_analysis}.

We categorize the observed visual artifacts into three primary failure modes: (i) \textit{Structural hallucination:} In the RoboCasa environment, the model incorrectly synthesizes a duplicate handle on the cup. (ii) \textit{Multi-view inconsistency:} In the LIBERO setup, a mug placed on a plate is successfully rendered in the primary view but disappears in the wrist-mounted view. (iii) \textit{Fine-grained omission:} In the real-world experiment, the model struggles to render fine-grained components, resulting in a missing door handle.

Remarkably, these failure cases motivate our core hypothesis: pixel-space representations are inherently sensitive to hallucinations, whereas latent-space dynamics provide a more robust foundation for policy learning. Even when the generated pixels exhibit localized artifacts, the underlying temporal dynamics and physical interactions remain accurate. This supports our design choice to leverage latent dynamics, which capture the essential task progression without being bottlenecked by the need for pixel-perfect generation.

\subsection{Visual Artifacts Corrupt IDM Labels}
\label{sec:idm_artifacts}

DreamGen~\cite{jang2025dreamgen} supervises VLAs with pseudo-actions inferred by a frozen IDM from decoded WM rollouts. We study this pixel bottleneck by running the same frozen IDM on 50 held-out LIBERO-Object scenes, changing only the input video source: ground-truth videos versus decoded WM rollouts. We then compute action error against temporally aligned ground-truth actions. Table~\ref{tab:visual-artifacts} shows that decoded rollouts produce noisier pseudo-labels under the same IDM.

\begin{table}[!ht]
    \centering
    \caption{\textbf{Frozen-IDM action error on LIBERO-Object.}
    Errors are computed against temporally aligned ground-truth actions over 50 held-out scenes. Decoded WM rollouts increase both MSE and MAE under the same IDM.}
    \label{tab:visual-artifacts}
    \begin{tabular}{@{}lcc@{}}
        \toprule
        \textbf{IDM input video} & \textbf{MSE} $\downarrow$ & \textbf{MAE} $\downarrow$ \\
        \midrule
        Ground-truth video    & 0.082 & 0.167 \\
        Decoded WM rollout    & 0.097 & 0.199 \\
        \midrule
        \textit{WM / ground-truth} & $1.18{\times}$ & $1.19{\times}$ \\
        \bottomrule
    \end{tabular}
\end{table}

Replacing real videos with decoded WM rollouts raises pseudo-action MSE by $18\%$ and MAE by $19\%$. Since the IDM is fixed and evaluation is performed on temporally aligned action sequences, the extra error enters through the decoded rollout stream rather than the pseudo-labeling model. This gives a concrete failure path for the LIBERO drop in Table~\ref{tab:libero_results_compact}: pixel-space WM supervision can turn visual artifacts into incorrect action targets. \textsc{World2Act} avoids this bottleneck by aligning policy actions to WM dynamics in latent space, without decoding frames for IDM supervision.
\subsection{Failure of WM to VLA Transfer}
\label{appendix-subsec:wm-vla-transfer-failure}

\begin{figure}[!ht]
    \centering
    \begin{subfigure}[b]{0.48\linewidth}
        \centering
        \includegraphics[width=\linewidth]{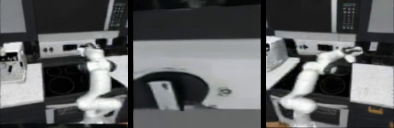}
        \caption{WM Imagination}
        \label{fig:wm_imagination}
    \end{subfigure}
    \hfill
    \begin{subfigure}[b]{0.48\linewidth}
        \centering
        \includegraphics[width=\linewidth]{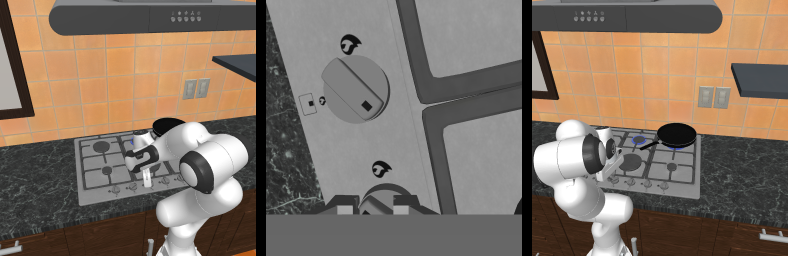}
        \caption{VLA Execution}
        \label{fig:vla_execution}
    \end{subfigure}

    \caption{\textbf{Failure analysis of WM to VLA transfer.} The WM generates a visually plausible completion of the task, but the VLA fails to physically ground these dynamics, causing a distribution shift during actual deployment.}
    \label{fig:transfer_failure}
\end{figure}

\autoref{fig:transfer_failure} illustrates a failure case when transferring from WM imagination to execution using GR00T-N1.6-ft+\textsc{World2Act}. We show the final frames of both the imagined and executed sequences for a \textit{Turn Off Stove} task. In the imagination sequence, the WM successfully predicts the robot grasping and turning the stove knob. However, the VLA fails to secure a firm grip on the knob during actual control, resulting in task failure. We attribute this discrepancy to the rigid physical constraints of robotic kinematics, which are far more complex and higher-dimensional than pixel space. While \textsc{World2Act} is specifically designed to bridge this gap by grounding the VLA in the WM's latent dynamics, this failure case highlights the difficulty of aligning visual imagination with precise low-level motor control.

\section{Broader Impact}
\label{appendix-sec:broader-impact}

\textsc{World2Act} provides a practical post-training framework for transferring world-model dynamics priors to vision-language-action policies. Instead of requiring pixel-space rollout supervision or changes to the underlying policy architecture, \textsc{World2Act} aligns policy action representations with latent dynamics induced by a pretrained world model. This makes the approach lightweight to apply to existing VLA systems and potentially useful as an additional post-training stage for robotic manipulation.

We will release our code and processed datasets to support reproducibility and further research in robot learning. In addition to training robot world models, these datasets provide fine-grained video-language demonstrations that may be useful for offline reinforcement learning, behavior cloning, action-conditioned video generation, and benchmarking short-horizon manipulation skills.

\section{Remarks on Related Work}
\label{appendix-sec:extra-related-work}

We further clarify how \textsc{World2Act} differs from the closest prior work that uses latent world-model or future-state representations to improve VLA performance:
\begin{itemize}
    \item \textbf{Cosmos Policy}~\cite{kim2026cosmos}. Cosmos Policy learns a joint embedding space and optimizes prediction heads for actions, future frames, and values. In contrast, \textsc{World2Act} does not train a new joint WM-policy architecture; it post-trains an existing VLA by aligning frozen-WM video dynamics with VLA action latents through lightweight video--action adapters and a residual action-latent policy.

    \item \textbf{V-JEPA 2}~\cite{assran2025v}. V-JEPA 2 performs model-based planning by steering the action distribution toward a desired final state using a latent-space objective over predicted future states. \textsc{World2Act} addresses a different setting: rather than optimizing actions toward a terminal state, it transfers the WM's temporally extended dynamics trajectory into the VLA action space via step-wise latent alignment, thereby guiding how the task should evolve over time.

    \item \textbf{FLARE}~\citep{zheng2025flare}. FLARE augments a diffusion/flow-matching VLA with learnable future tokens whose hidden states are aligned to embeddings of future observations, encouraging the policy to anticipate and steer toward desirable future states. While this provides an effective goal-oriented latent regularizer, it does not explicitly transfer a temporally dense dynamics trajectory from a frozen generative WM into the action space. In contrast, \textsc{World2Act} leverages WM-imagined latent dynamics as step-wise supervision targets and aligns them with VLA action latents through chunk-level video--action contrastive learning.

    \item \textbf{CoWVLA}~\citep{yang2026chain}. CoWVLA proposes a Chain-of-World VLA pretraining scheme that uses a pretrained video VAE to factor video segments into structure and motion latents, then jointly models sparse visual keyframes and action tokens with a unified autoregressive decoder. Although CoWVLA reduces redundant intermediate-frame prediction, it still requires co-training a large VLA backbone with keyframe and action-token modeling. In contrast, \textsc{World2Act} is a lightweight post-training method for existing VLAs: it keeps the backbone frozen, uses pre-decoding spatial video latents from a frozen WM as dynamics targets, and trains only video--action adapters plus a residual action-latent policy.
\end{itemize}

\end{document}